%% file: arxiv.tex
\PassOptionsToPackage{table}{xcolor}
\documentclass{article} % For LaTeX2e
\usepackage{iclr2027_conference,times}

\input{math_commands.tex}

\usepackage{hyperref}
\usepackage{url}
\usepackage{amssymb}
\usepackage{booktabs}
\usepackage{graphicx}
\usepackage{multirow}
\usepackage{caption}
\usepackage{subcaption}
\usepackage{amsmath}
\usepackage{algorithm}
\usepackage[noend]{algpseudocode}
\newcommand{\ind}{\mathbb{I}}
\newcommand{\vstart}{\texttt{<vstart>}}
\newcommand{\vend}{\texttt{<vend>}}

\DeclareMathOperator{\sg}{sg}
\DeclareMathOperator{\Norm}{Norm}
\DeclareMathOperator{\Std}{Std}
\usepackage[table]{xcolor}
\definecolor{strongred}{RGB}{255, 150, 150}
\definecolor{pinkred}{RGB}{241,144,140} 
\definecolor{lightgreen}{RGB}{226, 242, 226}
\definecolor{lightred}{RGB}{255, 225, 225}
\definecolor{stronggreen}{RGB}{102, 193, 140}
\title{Adaptive Look, Flexible Thought: Learning How and How Long to Think in Multimodal Reasoning}
\iclrfinalcopy
\author{%
  \begin{minipage}[t]{\dimexpr\textwidth-2\tabcolsep-15pt\relax}
    \raggedright
    {
      Quanlin Chen\textsuperscript{1,2},
      Xiuwei Chen\textsuperscript{1},
      Kun Xiang\textsuperscript{1},
      Zehua Ma\textsuperscript{1},
      Chunjin Ye\textsuperscript{1,3},
      Boqiang Guo\textsuperscript{4},
      Xianming Liu\textsuperscript{5,2*},
      Xiaodan Liang\textsuperscript{1*}\\[0.35em]
    }
    {\footnotesize\mdseries\upshape
      \textsuperscript{1} Sun Yat-sen University \\
      \textsuperscript{2} Shenzhen Loop Area Institute \\
      \textsuperscript{3} Beijing Zhongguancun Academy \\
      \textsuperscript{4} Huawei Technologies Co., Ltd\\
      \textsuperscript{5} Harbin Institute of Technology \\
    }
    \mdseries\itshape chenqlin7@mail2.sysu.edu.cn
  \end{minipage}%
}

\begin{document}
\begingroup
\renewcommand{\thefootnote}{\fnsymbol{footnote}}
\footnotetext[1]{Corresponding author}
\endgroup
\maketitle
\lhead{}  % 清除会议页眉
\renewcommand{\headrulewidth}{0pt}
\begin{abstract}
The introduction of long chain-of-thought (CoT) reasoning and external tools has substantially improved the performance of multimodal large language models (MLLMs) on general and reasoning tasks. However, existing fast-slow reasoning paradigms mainly regulate textual reasoning length, while visual computation introduced by external tools or visual tokens is often treated as an auxiliary component. This separation prevents MLLMs from jointly optimizing visual perception and textual reasoning.
To address this issue, we propose \textbf{L2T} (adaptive \textbf{L}ook, flexible \textbf{T}hought), a unified reasoning framework that jointly optimizes the \textbf{form} and \textbf{depth} of computation. L2T first discretizes tool-produced visual states into visual tokens, bringing visual perception and textual reasoning into a shared autoregressive space. Based on this unified representation, L2T introduces a modality-aware routing reward to select appropriate reasoning modes and a difficulty-aware fast-slow strategy to adapt reasoning depth according to task complexity.
Experiments on general perception and reasoning-intensive tasks demonstrate the effectiveness of \textbf{L2T}. Compared with explicit tool-calling approaches, \textbf{L2T} maintains competitive performance while achieving 1.7--3.6$\times$ higher throughput.
\end{abstract}

\section{Introduction}

Multimodal large language models (MLLMs) have demonstrated strong capabilities in visual perception and reasoning. However, their performance gains often rely on increasing test-time computation, typically through long Chain-of-Thought (CoT) reasoning or external tool use~\citep{Xu_2025_ICCV,huang2026visionr,wang2026perceptionaware}. While effective, these approaches introduce two key limitations: %(i) computation is allocated uniformly across samples regardless of difficulty
(i) they lack explicit mechanisms to adapt computation budgets according to sample difficulty, and (ii) it is expressed almost entirely in text, which becomes unreliable when answers hinge on fine-grained visual evidence~\citep{cao2025groundr1incentivizinggroundedvisual,tian2025thoughtaccuracydualnature}.

To address these issues, the `Thinking with Images' paradigm first introduces external visual tools such as cropping, zooming, and transformation to improve visual grounding~\citep{Wu_2024_CVPR,openai2025thinkingwithimages,hong2025deepeyesv2agenticmultimodalmodel,zhang2025thyme}. Although this improves reliability, it incurs significant overhead due to code generation, tool execution, and image feedback. Recent latent reasoning methods alleviate part of this latency by replacing explicit tool use with continuous latent representations~\citep{yang2025machinementalimageryempower,ray2025mulltokensmodalityagnosticlatentthinking,wang2025monetreasoninglatentvisual,chen2026latentvisualstatesefficient}. However, they typically rely on fixed-length latent sequences and cannot dynamically adapt computation budgets to task difficulty or visual complexity.

These limitations raise a fundamental question: how should an MLLM allocate computation during multimodal reasoning? We argue that efficient multimodal reasoning is not only about \emph{how much} to think, but also about \emph{what form} the computation should take. Perception-heavy tasks require fine-grained processing of local regions, spatial relations, and visual attributes, where purely textual rationales may drift from image evidence and rely on language priors~\citep{tian2025thoughtaccuracydualnature}. In contrast, reasoning-heavy tasks benefit from explicit symbolic abstraction and multi-step textual composition. Therefore, an efficient model should dynamically route computation between visual and textual modalities while also adjusting its reasoning depth according to task difficulty.
\begin{figure*}[htbp]
    \centering
    % \vspace*{-2.7em}
    \includegraphics[width=1.0\linewidth]{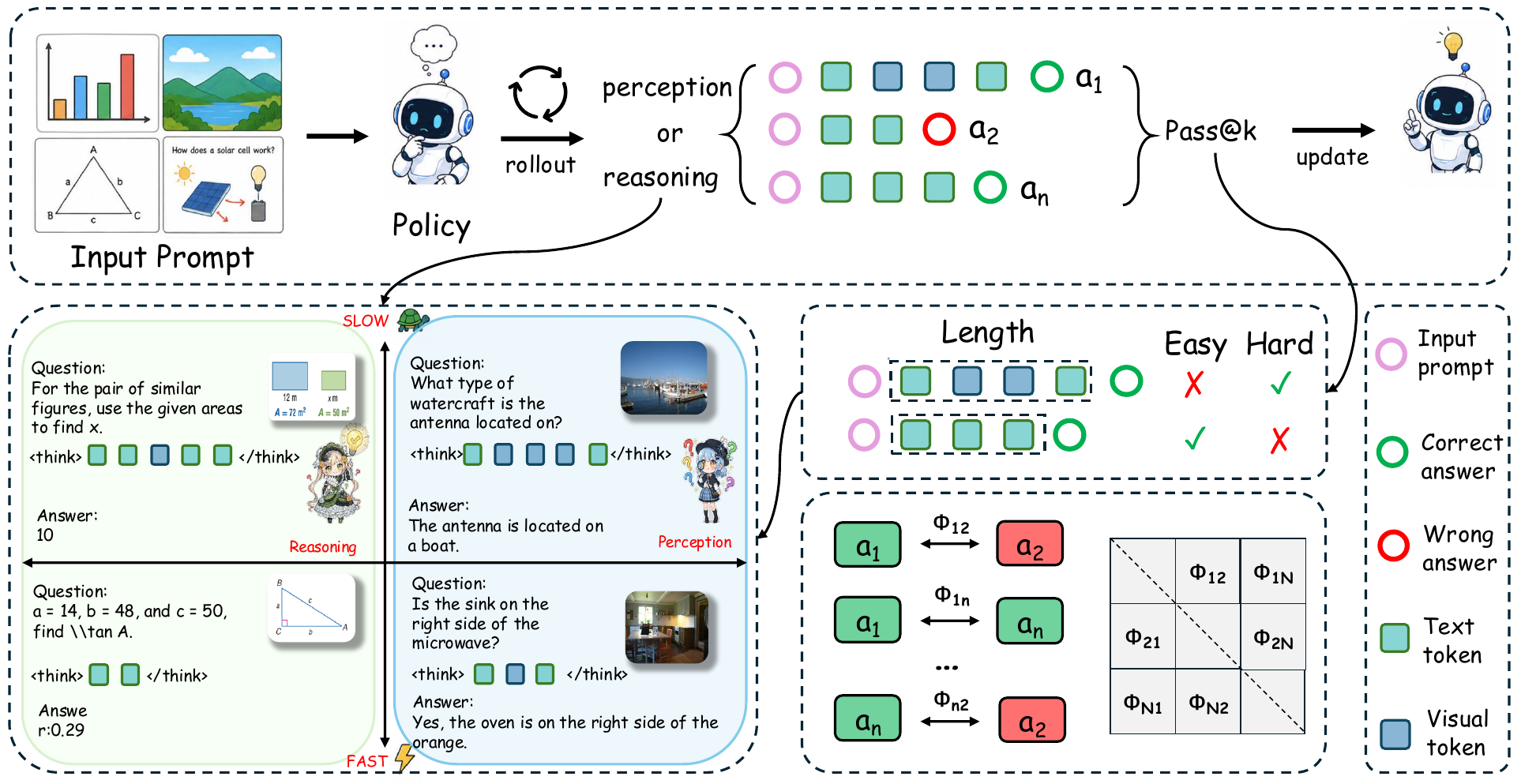}
    \caption{
    Overview of modality-aware routing and difficulty-aware fast-slow optimization.
    }
    \label{fig:fig0_main}
\end{figure*}

 Our motivation is shown in Figure~\ref{fig:fig0_main}, we propose an adaptive computation allocation framework for MLLMs. We first discretize tool-produced intermediate visual states into visual tokens, allowing visual evidence that would otherwise require external tool feedback to be represented within a single autoregressive sequence. Built on this representation, we introduce two adaptive training signals: a modality-aware routing reward that allocates computation between visual and textual tokens, and a difficulty-aware fast-slow advantage mechanism that adjusts reasoning length according to sample difficulty. Together, these components allow the model to decide both \emph{how} to think and \emph{how long} to think.
Our contributions are summarized as follows:
\begin{itemize}
    \item[1)] We introduce an adaptive \textbf{L}ook, flexible \textbf{T}hought (L2T) framework that represents visual perception and textual reasoning within the same autoregressive token space, enabling unified optimization of both computation modality and computation budget for both perception and reasoning tasks.
    %\item[2)] We propose an adaptive fast-slow reasoning objective that jointly learns modality-aware routing and difficulty-aware reasoning budgets, allowing the model to choose both the form and amount of computation.
    \item[2)] Building upon this unified representation, we propose an adaptive fast-slow reasoning objective that jointly optimizes computation modality and reasoning depth through modality-aware routing and difficulty-aware budget allocation, enabling the model to determine how and how long to think.
    %\item[3)] Experiments show that our method improves token efficiency and inference throughput while maintaining strong multimodal reasoning performance, supporting adaptive computation rather than uniform overthinking.
    \item[3)] Experiments demonstrate that L2T improves inference efficiency and token utilization while preserving strong multimodal reasoning performance across perception, reasoning, and OOD benchmarks.
\end{itemize}
\section{Related Work}

\subsection{Thinking about Images}

MLLMs have rapidly advanced visual understanding and multimodal reasoning by aligning visual representations with large language models
~\citep{bai2025qwen25vltechnicalreport}.
Following the success of CoT prompting in LLMs, many MLLM methods improve performance by encouraging longer textual rationales, self-reflection, or additional test-time computation
~\citep{lightman2024lets,Xu_2025_ICCV,11347549}.
However, thinking about images purely through text can be unreliable.
Recent studies show that textual CoT may be insufficient or even harmful for visual reasoning, especially when tasks require fine-grained perception, spatial grounding, or faithful use of image evidence
~\citep{xiao2025fastslow,tian2025thoughtaccuracydualnature}.
As textual reasoning becomes longer, models may increasingly rely on language priors rather than visual clues, leading to degraded performance and visual hallucination.
These findings suggest that stronger multimodal reasoning should not be reduced to simply generating more textual tokens.

\subsection{Thinking with Images}

To overcome the limitations of text-centric reasoning, recent `Thinking with Images' methods incorporate visual operations or intermediate visual states into the reasoning process ~\citep{openai2025thinkingwithimages}. Tool-based approaches augment MLLMs with visual operations such as cropping, zooming, grounding, and other image manipulations ~\citep{Wu_2024_CVPR,cao2025groundr1incentivizinggroundedvisual}. Recent systems, including PyVision, Thyme, and DeepEye v2, further extend visual reasoning through executable tool use ~\citep{zhao2025pyvision,zhang2025thyme,hong2025deepeyesv2agenticmultimodalmodel}. Despite their effectiveness, these methods rely on code generation, tool execution, and image feedback, resulting in substantial token overhead and latency.

An alternative direction internalizes visual computation through latent reasoning, replacing explicit reasoning chains with continuous latent embeddings. Recent MLLMs extend this paradigm to visual reasoning by aligning latent states with intermediate visual representations ~\citep{yang2025machinementalimageryempower,ray2025mulltokensmodalityagnosticlatentthinking,chen2026latentvisualstatesefficient}. However, these methods typically employ continuous representations with a fixed number of latent tokens, limiting their adaptability and preventing token-level optimization with reinforcement learning ~\citep{ray2025mulltokensmodalityagnosticlatentthinking,wang2025monetreasoninglatentvisual}. In contrast, our method discretizes intermediate visual computation into tokenizer-level visual tokens, enabling unified autoregressive generation and token-level reinforcement learning.

\subsection{Fast-Slow Thinking}
Although increasing reasoning depth generally improves performance, it also incurs substantially higher inference costs due to the quadratic complexity of LLMs and MLLMs with respect to sequence length. Recent approaches have alleviated insufficient reasoning by extending CoT or incorporating external tools~\citep{huang2026visionr,11347549,hong2025deepeyesv2agenticmultimodalmodel}. However, uniformly increasing the reasoning budget often leads to overthinking on simple problems. Besides wasting computational resources, excessive reasoning may even degrade accuracy~\citep{liu-etal-2026-think-faster,li2026signaloppositemeaningdirectioninformed}. This issue is particularly pronounced in MLLMs, where overly long reasoning traces can gradually weaken attention to the input image, leading to visual hallucinations~\citep{tian2025thoughtaccuracydualnature}. These observations have motivated recent research on fast-slow reasoning, which aims to allocate computation more adaptively. Kimi~\citep{team2025kimi} introduce length-aware objectives to encourage concise reasoning, while TWYN~\citep{yang2025thinkneedselfadaptivechainofthought}, DAST~\citep{shen-etal-2025-dast}, CosineReward~\citep{pmlr-v267-yang25ae}, and CARE~\citep{Wang_2026_CVPR} explicitly favor shorter yet correct reasoning traces. FAST further incorporates both image complexity and question difficulty to adapt reasoning length. DeepEyes v2 identifies samples that benefit from tool use through rule-based filtering and reinforcement learning, enabling adaptive tool invocation~\citep{hong2025deepeyesv2agenticmultimodalmodel}. ATLAS introduces a regularization objective to discourage the overuse of latent tokens~\citep{guo2026atlasagenticlatentvisual}. Despite these advances, existing methods do not provide a unified framework that jointly allocates textual and visual reasoning according to task requirements.
% TODO: This subsection will discuss adaptive test-time computation,
% difficulty-aware reasoning length, and fast-vs-slow reasoning.
% We will position our method as jointly learning when to think,
% how long to think, and whether to think with visual or textual tokens.

\section{Method}

%\td{Remove?}
\subsection{Discrete Visual Tokenization}

Inspired by previous works~\citep{peng2022beitv2maskedimage,11071901}, we introduce the Vector-Quantized Visual Tokenizer (VQ-VT) to bridge visual computation and autoregressive language modeling through a unified token space. Specifically, VQ-VT converts tool-produced intermediate visual states into discrete latent visual tokens that can be generated and consumed alongside textual tokens within the same autoregressive process. This unified representation enables visual and textual computation to be allocated under a common token budget, providing the foundation for our modality-aware routing and adaptive computation allocation. Following previous studies ~\citep{peng2022beitv2maskedimage,11209098}, we perform reconstruction in the feature space rather than the pixel space, as feature-level reconstruction has been shown to preserve semantic information more effectively.

\paragraph{Visual Codebook.}
Following VQ-VAE~\citep{NIPS2017_7a98af17}, we build a visual
tokenizer on top of a frozen vision encoder. Given an image, the frozen
encoder extracts patch features $\{h_i\}_{i=1}^{N}$, which are mapped
by a trainable linear encoder to latent features $\{z_i\}_{i=1}^{N}$.
We $\ell_2$-normalize both the latent features and the codebook
embeddings $\mathcal{V}=\{e_j\}_{j=1}^{K}$, and perform codebook lookup as
\begin{equation}
k_i=\arg\max_{1\leq j\leq K} z_i^\top e_j,
\qquad q_i=e_{k_i}.
\end{equation}
%A trainable linear decoder $D_\phi$ reconstructs the original vision feature from $q_i$. Using the standard straight-through estimator for the hard lookup, we optimize
%\begin{equation}
%\begin{aligned}
%\mathcal{L}_{\mathrm{VQ}}
%=
%\frac{1}{N}\sum_{i=1}^{N}
%\left[
%\|h_i-D_\phi(q_i)\|_2^2\\
%+\|\mathrm{sg}(z_i)-q_i\|_2^2
%+\beta\|z_i-\mathrm{sg}(q_i)\|_2^2
%\right],
%\end{aligned}
%\end{equation}
A trainable linear decoder $D_\phi$ reconstructs the original vision feature from
the quantized feature $q_i$, which is obtained by mapping $z_i$ to its nearest
neighbor in a codebook of size $K$. Using the standard straight-through estimator
for the hard lookup, we optimize

\begin{equation}
\begin{aligned}
\mathcal{L}_{\mathrm{VQ}}
&=
\frac{1}{N}\sum_{i=1}^{N}
\biggl[
\|h_i-D_\phi(q_i)\|_2^2 \\
&\quad +\|\mathrm{sg}(z_i)-q_i\|_2^2
+\alpha\|z_i-\mathrm{sg}(q_i)\|_2^2
\biggr],
\end{aligned}
\end{equation}

%where $\mathrm{sg}$ denotes stop-gradient. After training, the $K$ code indices and two boundary markers are added to the MLLM vocabulary as discrete visual tokens.
where $\mathrm{sg}$ denotes the stop-gradient operator, and $\alpha$ is a hyperparameter balancing the training of the codebook and the encoder. After training, the $K$ codebook indices and two boundary markers are added to the MLLM vocabulary as discrete visual tokens, which enable visual and textual reasoning to share a unified autoregressive token space for adaptive reasoning depth.

\subsection{SFT Dataset Construction}

We transform existing tool-calling data into discrete visual reasoning trajectories.

Given a sample $(X,I,C,A)$, where $X$ is the query, $I$ is the input image, $C$ denotes tool-calling code, and $A$ is the answer, we execute $C$ on the input image $I$ to obtain intermediate visual states, the intermediate image is encoded by the pre-trained VQ-VT encoder and quantizer:

\begin{gather}
I'=\text{Execute}(C,I).\\
V
=
\mathcal{Q}(\mathcal{E}(I'))
=
\{v_{1},\ldots,v_{m}\},~v_{j}\in\mathcal V
\end{gather}

Finally, we replace the original tool code with the corresponding visual token sequences:
\begin{equation}
D=
\{
X,
text,
v_{start},
v_1,v_2,\dots,v_n,
v_{end},
text,
A
\}.
\end{equation}

\subsection{Three-Stage Training Strategy.}

\paragraph{Stage 1: Learning to Think Visually.}

We first perform supervised fine-tuning on the constructed dataset to initialize visual reasoning. Since visual reasoning trajectories are represented as discrete visual tokens, the model learns to generate these tokens at appropriate reasoning steps via standard next-token prediction:

\begin{equation}
\mathcal L_{\mathrm{SFT}}
=
-\sum_i
\log
P(s_i|s_{<i},X,I).
\end{equation}

After this stage, the model acquires the ability to invoke visual tokens and associate them with corresponding visual transformations.
\paragraph{Stage 2: Learning to Think Reliably.}

Stage 2 performs reinforcement learning to establish a stable reasoning policy before introducing adaptive routing. Each response is required to follow

\begin{equation}
y=
\texttt{<think>}~z~\texttt{</think>}
\;
\texttt{<answer>}~\hat a~\texttt{</answer>},
\end{equation}

where the \texttt{<think>} field contains the reasoning process and the \texttt{<answer>} field is used for answer verification.

The Stage 2 reward is defined as

\begin{equation}
r_{\mathrm{stable}}
=
r_{\mathrm{acc}}
\cdot
r_{\mathrm{fmt}},
\end{equation}

where

\begin{equation}
r_{\mathrm{acc}}
=
\mathbb I
[
\mathrm{Verify}
(\hat a,a^\ast)
],
~r_{\mathrm{fmt}}
=
\mathbb I
[
y\in\mathcal F
].
\end{equation}

Here, $a^\ast$ is the correct answer, $\mathcal F$ denotes the set of valid responses following the required
\texttt{<think>}--\texttt{<answer>} format.

This warm-up stage encourages the model to generate correct and well-structured responses, providing a stable initialization for the adaptive routing and fast-slow reasoning objectives introduced in Stage 3.

\paragraph{Stage 3: Learning to Think Adaptively.}
As shown in Figure~\ref{fig:fig0_main}, after learning to produce stable and verifiable responses, the model is further optimized to reason adaptively. Specifically, Stage 3 jointly learns (1) modality-aware routing between visual and textual reasoning and (2) difficulty-aware reasoning budgets.
This stage turns stable reasoning into adaptive reasoning. It has two coupled goals: selecting the appropriate reasoning modality for each task and allocating a suitable reasoning token for each sample, which help model learns
to allocate computational resources more selectively according to task demands. %The first goal is handled by modality-aware routing, and the second by a pass-rate-based fast-slow advantage objective.
\paragraph{Modality-aware Routing.} To learn adaptive reasoning modality, we construct a modality-aware Routing reward.
Given the reasoning trace $z$ within the \texttt{<think>} field, let $V(z)$ and $T(z)$ denote the numbers of visual and textual reasoning tokens. Their normalized ratios are

\begin{equation}
\rho_v(z)=\frac{V(z)}{V(z)+T(z)},
\qquad
\rho_t(z)=\frac{T(z)}{V(z)+T(z)}.
\end{equation}

The visual and textual activation scores are defined as

\begin{align}
s_v&=
\min\!\left(\frac{\rho_v}{\tau_v},1\right)
\min\!\left(\frac{V}{V_{\min}},1\right),\\
s_t&=
\min\!\left(\frac{\rho_t}{\tau_t},1\right)
\min\!\left(\frac{T}{T_{\min}},1\right),\\
s_v^{\mathrm{mix}}&=
\min\!\left(\frac{\rho_v}{\tau_v^{\mathrm{mix}}},1\right)
\min\!\left(\frac{V}{V_{\min}},1\right),\\
s_t^{\mathrm{mix}}&=
\min\!\left(\frac{\rho_t}{\tau_t^{\mathrm{mix}}},1\right)
\min\!\left(\frac{T}{T_{\min}},1\right).
\end{align}
For each training sample, we assign two binary routing labels.
$n_v,n_t\in\{0,1\}$,
where $n_v$ and $n_t$ indicate whether visual-centered and text-centered reasoning are required, respectively. Routing labels are initialized based on dataset/task categories and further refined through keyword-based rules.

The route score is computed according to four routing types:

\begin{equation}
\tilde{s}_{\mathrm{route}}=
\begin{cases}
s_v,
& n_v=1,n_t=0,\\
s_t\lambda_{\mathrm{novis}}^{\mathbb I[V>0]},
& n_v=0,n_t=1,\\
\min(s_v^{\mathrm{mix}},s_t^{\mathrm{mix}}),
& n_v=1,n_t=1,\\
\lambda_{\mathrm{simple}}^{\mathbb I[V>0]},
& n_v=0,n_t=0.
\end{cases}
\end{equation}
The visual-only route rewards visual-token usage through $s_v$, which increases when the visual ratio $\rho_v$ and the number of visual tokens $V$ are sufficiently large. The text-only route rewards textual reasoning through $s_t$, but additionally applies the penalty $\lambda_{\mathrm{novis}}$ whenever any visual token is generated. Moreover, when $T$ is fixed, increasing $V$ lowers $\rho_t=T/(T+V)$, so the text-only score can decrease further as more visual tokens are used. In contrast, the simple route only applies a binary penalty $\lambda_{\mathrm{simple}}$ once $V>0$; additional visual tokens do not cause extra penalty beyond this indicator. The mixed route requires both visual and textual channels to be active by taking the minimum of the two mixed activation scores.

The route score is incorporated as a soft shaping reward,

\begin{equation}
R_i
=
\left(0.9\cdot r_{\mathrm{acc},i}+0.1\cdot s_{\mathrm{route},i}\right)
\cdot r_{\mathrm{stable}},
\end{equation}

Thus, correctness remains the primary optimization target, and the route score only acts as a weak shaping signal among valid responses. This prevents routing from overriding answer correctness while still encouraging the model to express computation in the modality required by the task.

\paragraph{Difficulty-aware Fast-Slow Thinking.}

To learn adaptive reasoning budgets, we construct a pass-rate-based contrastive objective.

For a rollout group $G$, the group pass rate is

\begin{equation}
q_i=\mathbb I[R_i>\tau_{\rm corr}],
\qquad
p_G=\frac1{|G|}\sum_{i\in G}q_i .
\end{equation}
Here, $\tau_{\rm corr}$ represents the threshold score for correctness. For simplicity, we set $\tau_{\rm corr}\approx1$, which requires both the route reward and the final answer to be correct.

Given rollout reward $R_i$ and reasoning length $L_i$, we define

\begin{equation}
\Delta R_{ij}=R_i-R_j,
\qquad
\Delta L_{ij}
=
\tanh\!\left(
\frac{L_i-L_j}{\max(\sigma_L,L_{\min})}
\right).
\end{equation}
Here, $\sigma_L$ denotes the standard deviation of reasoning lengths within the rollout group. The thresholds $\tau_{\rm easy}$ and $\tau_{\rm hard}$ divide prompts into easy, medium, and hard groups according to the group pass rate.

The pairwise potential is

\begin{align}
c_{ij}
&=
\mathbb I[R_i>\tau_{\rm corr}]
\mathbb I[R_j>\tau_{\rm corr}],\\
h_{ij}
&=
\mathbb I[R_i>\tau_{\rm corr}]
\mathbb I[R_j\le \tau_{\rm corr}].
\end{align}

\begin{equation}
\Phi_{ij}
=
\begin{cases}
\Delta R_{ij}
-
\beta\Delta L_{ij}c_{ij},
&
p_G\ge\tau_{\mathrm{easy}},\\[2pt]
\Delta R_{ij},
&
\tau_{\mathrm{hard}}
\le
p_G
<
\tau_{\mathrm{easy}},\\[2pt]
\Delta R_{ij}
+
\beta'[\Delta L_{ij}]_+
h_{ij},
&
p_G<\tau_{\mathrm{hard}}.
\end{cases}
\end{equation}

This objective favors shorter successful traces on easy samples, preserves longer successful traces on hard samples, and uses only reward ranking for medium-difficulty samples. In this way, the model is not uniformly compressed; instead, its reasoning length is adapted according to empirical sample difficulty.

The sequence-level advantage is computed by aggregating pairwise potentials,

\begin{equation}
A_i
=
\mathrm{Norm}
\left(
\frac1{|G|-1}
\sum_{j\neq i}
\Phi_{ij}
\right),
\end{equation}

where $\mathrm{Norm}(\cdot)$ denotes group-wise normalization.

\paragraph{Policy Optimization.}
Finally, we optimize the policy using GRPO~\citep{Guo_2025} with our
dynamic advantage $A_i$:
\begin{equation}
\begin{aligned}
\mathcal L_{\mathrm{GRPO}}
&=
-\frac1{|G|}\sum_{i\in G}\frac1{|y_i|}
 \sum_t g_{i,t}
+\lambda_{\mathrm{KL}}
 D_{\mathrm{KL}}(\pi_\theta\|\pi_{\mathrm{ref}}),
\\[-2pt]
g_{i,t}
&=
\min\!\left\{
\rho_{i,t}A_i,\,
\operatorname{clip}(\rho_{i,t},1-\epsilon,1+\epsilon)A_i
\right\},
\\[-2pt]
\rho_{i,t}
&=
\frac{\pi_\theta(y_{i,t}\mid x,y_{i,<t})}
{\pi_{\theta_{\mathrm{old}}}(y_{i,t}\mid x,y_{i,<t})}.
\end{aligned}
\label{eq:grpo}
\end{equation}
Here, $y_i$ is a rollout in group $G$, $\pi_{\mathrm{ref}}$ is the
reference policy, and $A_i$ is computed from the pass-rate contrastive
objective. By encoding routing and length preferences in $A_i$, the
objective favors correct responses with appropriate routing and
deliberation length. The pseudo code is shown in Appendix~\ref{sec:pseudo}.
\section{Experiment and Discussion}
\subsection{Experimental Setup}
We conduct experiments based on Qwen2.5-VL~\citep{bai2025qwen25vltechnicalreport}. Experiments cover visual-centric perception and reasoning-centric understanding.
We compare our method with a diverse set of baselines, including general open-source models, textual reasoning methods, tool-calling and latent reasoning methods, as well as vanilla slow thinking RLVR baselines trained from the same SFT model. The slow-thinking RLVR baseline is trained under the same settings as our method and is primarily used to evaluate whether our fast-slow thinking framework effectively reduces the reasoning budget.
Detailed training configurations, dataset processing, and evaluation protocols are provided in the Appendix~\ref{detail}. Further experiments on  on the out-of-distribution (OOD) benchmarks are reported in Appendix~\ref{sec:additional}. Additional experiments on visual tokens are presented in Appendix~\ref{sec:visual}. More experiments on fast–slow thinking and CoT are provided in Appendix~\ref{sec:fast} and~\ref{sec:long-cot-effectiveness}, respectively.

%We compare our method with a diverse set of baselines, including general open-source models (Qwen2.5-VL~\citep{bai2025qwen25vltechnicalreport}), textual reasoning methods (Vision-R1~\citep{huang2026visionr}, FAST~\citep{xiao2025fastslow}, PAPO~\citep{wang2026perceptionaware}, and CARE~\citep{Wang_2026_CVPR}), tool-calling and latent reasoning methods (Thyme-RL~\citep{zhang2025thyme}, Monet~\citep{wang2025monetreasoninglatentvisual}, EVA~\citep{chen2026latentvisualstatesefficient}, and DeepEyesV2~\citep{hong2025deepeyesv2agenticmultimodalmodel}), as well as slow thinking RLVR baselines (vanilla GRPO~\citep{Guo_2025} and GSPO~\citep{zheng2025groupsequencepolicyoptimization}) trained from the same SFT model. The slow-thinking RLVR baseline is trained under the same settings as our method and is primarily used to evaluate whether our fast-slow thinking framework effectively reduces the reasoning budget.

\subsection{Main Results}
\begin{figure*}[htbp]
    \centering

    \begin{subfigure}[t]{0.42\textwidth}
        \centering
        \includegraphics[
            width=\linewidth,
            height=3.4cm,
            keepaspectratio
        ]{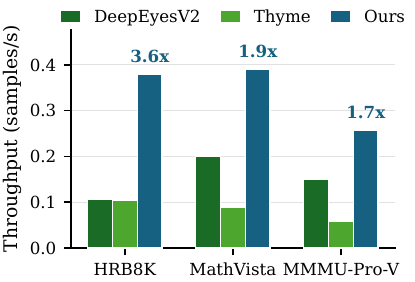}
        \caption{The comparison of throughput between tool-calling method and our method}
        \label{fig:throughput}
    \end{subfigure}
    \hfill
    \begin{subfigure}[t]{0.57\textwidth}
        \centering
        \includegraphics[
            width=\linewidth,
            height=3.4cm,
            keepaspectratio
        ]{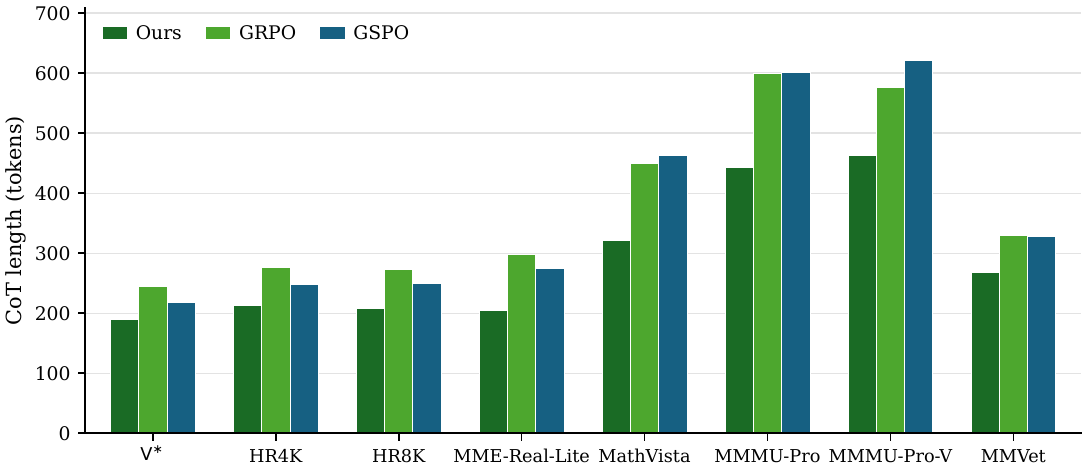}
        \caption{COT length comparison of RLVR baseline and our method}
        \label{fig:cot_length}
    \end{subfigure}

    \caption{Throughput and generation length comparison.}
    \label{fig:main_performance}
\end{figure*}
\begin{table*}[ht!]
\centering
\setlength{\tabcolsep}{3pt}
\small
\caption{
Performance comparison across perception and reasoningcentric multimodal benchmarks.
The best result is shown in \textbf{bold}, and the second-best result is \underline{underlined}. For benchmark results reported in prior papers, we directly cite the published numbers.For missing entries, we evaluate the corresponding models under the same environment and mark these results with $^\dag$. 
}
\resizebox{\textwidth}{!}{
\begin{tabular}{lccccccccc}
\toprule
\multirow{2}{*}{Models}
& \multicolumn{4}{c}{Visual-Centric Perception}
& \multicolumn{4}{c}{Reasoning-Centric Understanding} \\
\cmidrule(lr){2-5} \cmidrule(lr){6-9}
& V$^\ast$
& HR4K
& HR8K
& %MME-Real-Lite 
MME-Lite
& MMMU-Pro
& MMMU-Pro-V
& MathVista
& MMVet
& AVG\\
\midrule
\multicolumn{10}{c}{\textit{Open-Source Models}} \\
\midrule
Qwen2.5-VL-3B~\citep{bai2025qwen25vltechnicalreport}        & 74.9$^\dag$ & 66.1$^\dag$ & 63.5$^\dag$ & 42.4 & 31.6 & 28.5$^\dag$ & 62.3 & 61.8 & 53.9 \\
Qwen2.5-VL-7B~\citep{bai2025qwen25vltechnicalreport}        & 76.4$^\dag$ & 68.4$^\dag$ & 64.6$^\dag$ & 44.9 & 38.3 & 35.6$^\dag$ & 68.2 & 67.1 & 57.9 \\
\midrule
\multicolumn{10}{c}{\textit{Textual Reasoning Models}} \\
\midrule
Vision-R1-7B~\citep{huang2026visionr}        & 80.6$^\dag$ & 75.6$^\dag$ & 70.5$^\dag$ & 51.5$^\dag$ & 38.5$^\dag$ & 35.3$^\dag$ & 72.1 & \textbf{76.4$^\dag$} & 62.6 \\
FAST-3B~\citep{xiao2025fastslow}        & 69.1 & 61.1 & 54.7 & 40.9 & 35.0 & 29.8 & 66.2 & 64.0 & 52.6\\
FAST-7B~\citep{xiao2025fastslow}        & 76.4$^\dag$  & 73.1$^\dag$& 66.5$^\dag$ & 48.3$^\dag$ & 40.2$^\dag$ & 37.1$^\dag$ & \underline{73.8} & 71.2 & 60.8 \\
%PAPO-D-3B        & 74.8$^\dag$ & 66.0$^\dag$ & 63.6$^\dag$ & 44.1$^\dag$ & 28.8 & 30.9$^\dag$ & 62.5 & 69.5$^\dag$ & 55.0 \\
PAPO-3B~\citep{wang2026perceptionaware}        & 75.9$^\dag$ & 65.3$^\dag$ & 63.7$^\dag$ & 43.6$^\dag$ & 27.1 & 29.1$^\dag$ & 61.4 & 69.7$^\dag$ & 54.5 \\
%PAPO-D-7B        & 78.0$^\dag$ & 71.8$^\dag$ & 67.8$^\dag$ & 47.8$^\dag$ & 36.3 & \textbf{39.1$^\dag$} & 67.5 & 74.8$^\dag$ & 60.4 \\
PAPO-7B~\citep{wang2026perceptionaware}        & 78.5$^\dag$ & 71.5$^\dag$ & 67.5$^\dag$ & 47.0$^\dag$ & 36.6 & \underline{38.5$^\dag$} & 69.5 & 73.9$^\dag$ & 60.4 \\
CARE-3B~\citep{Wang_2026_CVPR}        & -- & -- & -- & -- & 33.1 & 24.0 & 66.5 & -- & --\\
CARE-7B~\citep{Wang_2026_CVPR}        & -- & -- & -- & -- & 39.7 & 37.1 & \textbf{74.7} & -- & -- \\
\midrule
\multicolumn{10}{c}{\textit{tool-calling and Latent Reasoning Models}} \\
\midrule
Thyme-RL~\citep{zhang2025thyme}        & 82.2 & \underline{77.0} & 72.0 & 55.2 & 24.6$^\dag$ & 33.6$^\dag$ & 70.0 & 71.3$^\dag$ & 60.7 \\
Monet~\citep{wang2025monetreasoninglatentvisual}       & \underline{83.3} & 71.0 & 68.0 & 55.5 & 36.0$^\dag$ & 31.2$^\dag$ & 63.6$^\dag$ & 60.4$^\dag$ & 58.6 \\
EVA~\citep{chen2026latentvisualstatesefficient}       & 80.2  & 74.0 & 69.5 & 49.8 & -- & -- & -- & -- \\
DeepEyesV2~\citep{hong2025deepeyesv2agenticmultimodalmodel}    & 81.8 & \textbf{77.9} & \textbf{73.8} & 50.4 & 39.1$^\dag$ & 34.6$^\dag$ & 70.1 & 64.7$^\dag$ & 61.6 \\

\midrule
\multicolumn{10}{c}{\textit{Slow Thinking RLVR Baseline}} \\
\midrule
SFT-7B+GRPO~\citep{Guo_2025}       & 81.7 & 74.8 & 71.8 & \underline{53.8} & 39.2 & 36.5 & 69.9 & 75.6 & \underline{62.9} \\
SFT-7B+GSPO~\citep{zheng2025groupsequencepolicyoptimization}       & 79.1 & 76.0 & 69.6 & \textbf{54.2} & \underline{40.3} & 37.1 & 67.8 & \underline{75.9} & 62.5\\
\midrule
\multicolumn{10}{c}{\textit{Our Models}} \\
\midrule
L2T-3B (Ours)    & 73.8 & 68.6 & 67.0 & 49.6 & 33.9 & 30.8 & 64.5 & 75.2 & 57.9\\
L2T-7B (Ours)    & \textbf{83.8} & 75.2 & \underline{72.7} & 53.0 & \textbf{40.7} & 37.1 & 72.9 & 75.4 & \textbf{63.9} \\
\bottomrule
\end{tabular}}
\label{tab:main_results}
\end{table*}
\paragraph{Compare Efficiency with tool-calling Method.}
We compare the inference throughput of our method with tool-calling approaches on a single H100 GPU. Experiments are conducted on three benchmarks with different characteristics: HRBench8K (high-resolution perception), MathVista$_{\mathrm{mini}}$ (reasoning), and MMMU-Pro-V (mixed perception and reasoning). As shown in Figure~\ref{fig:throughput}, our method consistently achieves substantially higher throughput. In particular, it is 3.6$\times$ faster than the strongest tool-based baseline, DeepEyesV2, on HRBench8K, and provides 1.7--1.9$\times$ speedups on the other two benchmarks. These results demonstrate that our method significantly reduces the computational overhead introduced by explicit tool execution.
\paragraph{Compare Token Length with RLVR Baseline.}
We further compare the generated respond length with RLVR baselines. As shown in Figure~\ref{fig:cot_length} compared with both vanilla GRPO and GSPO, our method consistently produces fewer tokens while achieving better performance, demonstrating that it improves reasoning efficiency by reducing token consumption without sacrificing accuracy.

\paragraph{Results on Standard Benchmarks.}
Table~\ref{tab:main_results} summarizes the performance of our method on standard multimodal benchmarks spanning both perception- and reasoning-centric tasks, verifies that
the efficiency gains do not come from a collapse in task ac-
curacy. Our method achieves the best overall average performance, outperforming textual reasoning, tool-calling reasoning, latent reasoning, and RLVR baselines trained from the same SFT initialization. Notably, existing reasoning-oriented MLLMs generally exhibit a trade-off between perception and reasoning: textual reasoning models excel at reasoning but often degrade visual perception, particularly on high-resolution benchmarks, whereas tool-calling and latent reasoning models achieve stronger perception but limited reasoning gains. In contrast, our method effectively mitigates this trade-off, achieving the highest average score (63.9) while remaining competitive across both perception- and reasoning-centric benchmarks.

\subsection{Ablation Studies}
\paragraph{Ablation on Route Reward}
To evaluate the proposed modality-aware routing reward, we remove this objective and first compare token allocation. As shown in Table~\ref{tab:special-token-ratio}, the full model substantially shortens the average CoT on both visual-centric and reasoning-centric tasks. At the same time, it increases the visual-token ratio much more on visual-centric tasks than on reasoning-centric tasks, indicating that routing does not simply maximize or minimize visual-token usage but instead allocates computational resources more selectively according to task demands while preserving task accuracy.
\begin{table}[htbp]
  \centering
  \small
  \setlength{\tabcolsep}{3pt}
  %\renewcommand{\arraystretch}{0.75}
  %\small
   \caption{Ablation on Route Reward and Difficulty-aware fast-slow.}
   \resizebox{0.8\textwidth}{!}{
  \begin{tabular}{lccc}
  \toprule
  Category & standard (\%) & w/o route reward (\%) &  w/o fast-slow \\
  \midrule
  \multicolumn{4}{c}{Ration of Visual Token} \\
  \midrule
  Visual-Centric &  31.8 & 21.9 (-9.9) & -- \\
  Reasoning-Centric & 11.56 & 11.03 (-0.53) & -- \\
  \midrule
  \multicolumn{4}{c}{Average Length of CoT} \\
  \midrule
  Visual-Centric &  206.8  &  292.6 (+85.8) &  234.7 (+27.9)  \\
  Reasoning-Centric &  416.2  &  547.0 (+130.8) &  469.9 (+53.7)  \\
  \midrule
  \multicolumn{4}{c}{Accuracy} \\
  \midrule
  Standard Benchmark&  63.9 &   63.8 (-0.1)&   63.7 (-0.2)\\
  %Visual-Centric & 71.2  & 71.0 & +0.2 \\
  %Reasoning-Centric & 56.5  & 56.6 & -0.1  \\
  \bottomrule
  \end{tabular}}
  \label{tab:special-token-ratio}
  \end{table}
%At the same time, it increases the visual-token ratio much more on visual-centric tasks than on reasoning-centric tasks, indicating that routing does not simply minimize visual-token usage; instead, it reallocates visual tokens toward samples where visual evidence is more useful and reduces unnecessary textual overthinking. After this efficiency gain, the average task accuracy is essentially preserved.  Which indicating that the model learns to allocate computational resources more selectively according to task demands.

\paragraph{Ablation on Difficulty-aware Fast-Slow.} 
To verify the necessity of difficulty-aware fast-slow reasoning, we remove the pass-rate-based length preference and first examine generated length. Table~\ref{tab:special-token-ratio} shows that the full model uses shorter traces than the ablated variant on both visual-centric and reasoning-centric tasks. This indicates that pass-rate difficulty signals help reduce unnecessary reasoning, especially on samples where shorter successful traces are sufficient. After this length reduction, the result also shows that average accuracy is preserved with only small changes. Which indicates that we did not pursue shorter CoT at the expense of accuracy.

\begin{table}[htbp]
  \centering
  \small
  %\setlength{\tabcolsep}{3pt}
  %\renewcommand{\arraystretch}{0.75}
  %\small
  \caption{Ablation on visual token and two-stage rl.}
  \resizebox{0.8\textwidth}{!}{
  \begin{tabular}{lccc}
  \toprule
  Components & Visual-Centric & Reasoning-Centric  & Total $\Delta$ \\
   \midrule
  w/o visual token & 70.6  & 55.4 & -1.7 \\
  w/o two-stage rl & 69.3  & 56.1 &  -2.3 \\
  standard & 71.2  &  56.5 & -- \\
  \bottomrule
  \end{tabular}}
  \label{tab:ablation}
  \end{table}
\paragraph{Ablation on Visual Token}
To verify the effectiveness of the proposed Visual Tokens, we replace all Visual Tokens with identical, semantically meaningless <unk> tokens in stage 1. The training for Stage 2 and Stage 3 is the same as the standard method. As shown in Table~\ref{tab:ablation}, our method still outperforms the baseline by nearly one percentage point. This result indicates that the improvement is not simply a consequence of extending the CoT, but instead stems from the meaningful semantic information encoded by the learned Visual Tokens.
\paragraph{Ablation on Two-Stage RL} 
To evaluate the necessity of the proposed two-stage RL framework, we directly initialize Stage 3 RL from the SFT model without performing Stage 2 training, while keeping the number of RL steps identical to that of the standard Stage 2 + Stage 3 setting. As shown in Table~\ref{tab:ablation}, bypassing Stage 2 results in a performance degradation of 2.3 percentage points. This observation indicates that the model should first acquire reliable reasoning capabilities before learning to optimize token allocation efficiently.
\section{Conclusion}
In this paper, we presented a unified computation modality and budgets allocation framework for efficient multimodal reasoning. By utilizing VQ-VT to discretize intermediate visual tool outputs into visual tokens, our framework unifies visual and textual computation within the same autoregressive token space, eliminating the need for external tool execution during inference. Built upon this representation, a modality-aware routing objective determines which form of computation to use, while a difficulty-aware fast-slow objective determines how much computation to allocate according to task difficulty. Experiments demonstrate that our framework consistently improves computational efficiency, without sacrificing multimodal reasoning performance. These results suggest that efficient multimodal reasoning is fundamentally a computation allocation problem. Rather than uniformly increasing reasoning effort, an MLLM should learn both which reasoning modality to employ for different task types and how much computation to allocate for tasks of different difficulty. Future work should develop richer visual representations to enhance perceptual representation.

\bibliography{iclr2027_conference}
\bibliographystyle{iclr2027_conference}

\appendix
\section{Appendix}
\subsection{Additional Experiment}
\label{sec:additional}
\begin{figure}[htbp]
    \centering
    % \vspace*{-2.7em}
    \includegraphics[width=0.6\linewidth]{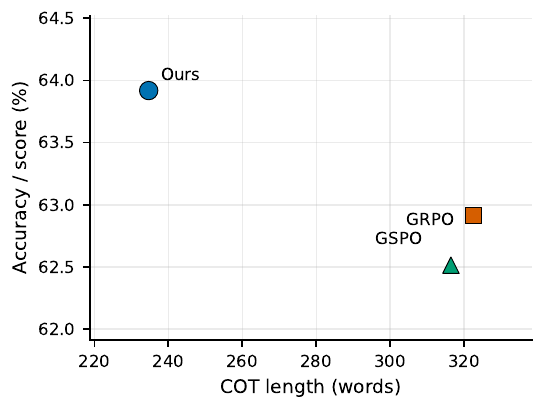}
    \caption{
        Pareto analysis of our method and slow-thinking RLVR Baseline.
    }
    \label{fig:pareto}
\end{figure}

\paragraph{Results on OOD Physics Reasoning}
To further assess the generalization ability of our method beyond the training distribution, we evaluate it on the out-of-distribution (OOD) SeePhys-family physics reasoning benchmarks, including SeePhys and SeePhysPro. As shown in Table~\ref{tab:seephys_results}, our method achieves the highest average score among the compared approaches, demonstrating its effectiveness in handling physics reasoning tasks under distribution shifts.  The consistent performance across the two benchmarks suggests that our method can improve the transferability of learned reasoning capabilities to unseen physics-related scenarios, rather than merely exploiting patterns in the training data. These results further highlight the potential of our approach to enhance the robustness of multimodal reasoning in OOD settings.
\begin{table}[htbp]
  \centering
  \setlength{\tabcolsep}{1.4pt}
  %\small
  %\renewcommand{\arraystretch}{1.0}
  \caption{
  Performance on SeePhys-family physics reasoning benchmarks.%Vision-only requires reading the full problem from the image. Level 1 removes the image and verbalizes visual information.
  }
  \begin{tabular}{lccccc}
  \toprule
  \multirow{2}{*}{Models}
  & \multicolumn{2}{c}{SeePhys}
  & \multicolumn{2}{c}{SeePhysPro}
  & \multirow{2}{*}{AVG} \\
  \cmidrule(lr){2-3} \cmidrule(lr){4-5}
  %\hline
  & Vanilla
  & Vision-only
  & Level 2
  & Level 1 
  & \\
  \midrule
  Qwen2.5-VL-7B        & 11.7 & 8.5 & 11.5 & 12.0 & 10.9 \\
 
  FAST-7B              & \textbf{13.4} & \textbf{10.1} & 11.5 & 13.0 & 12.0 \\
  %PAPO-D-7B              & \underline{14.4} & \textbf{10.4} & 11.0 & \underline{13.5} & 12.3 \\
  PAPO-7B              & 11.6 & 9.7 & 9.0 & 12.0 & 10.6 \\
  
  Thyme-RL         & 10.7 & 7.6 & 11.0 & 11.5 & 10.2 \\
  Monet              & 11.1 & 5.8 & \underline{15.0} & 12.5 & 11.1 \\
  DeepEyesV2              & 11.1 & 8.8 & 10.5 & 5.0 & 8.9 \\
  \midrule
  L2T-7B (Ours)              & \underline{12.9} & \underline{9.2} & \textbf{17.0} & \textbf{15.0} & \textbf{13.5} \\
  \bottomrule
  \end{tabular}

  \label{tab:seephys_results}
  \end{table}
\paragraph{Pareto analysis} To more intuitively illustrate the advantages of our method over the baselines, we present the results using a Pareto frontier analysis. As shown in Figure~\ref{fig:pareto}, compared with the slow-thinking baseline, our method achieves higher accuracy while consuming fewer tokens, demonstrating a more favorable trade-off between reasoning performance and computational efficiency. This suggests that our method can produce more accurate results with less reasoning overhead, improving token efficiency without sacrificing—and potentially even enhancing—reasoning quality.
\subsection{Analysis of VQ Visual Token}
\label{sec:visual}
\begin{figure}[htbp]
    \centering
    % \vspace*{-2.7em}
    \includegraphics[width=0.6\linewidth]{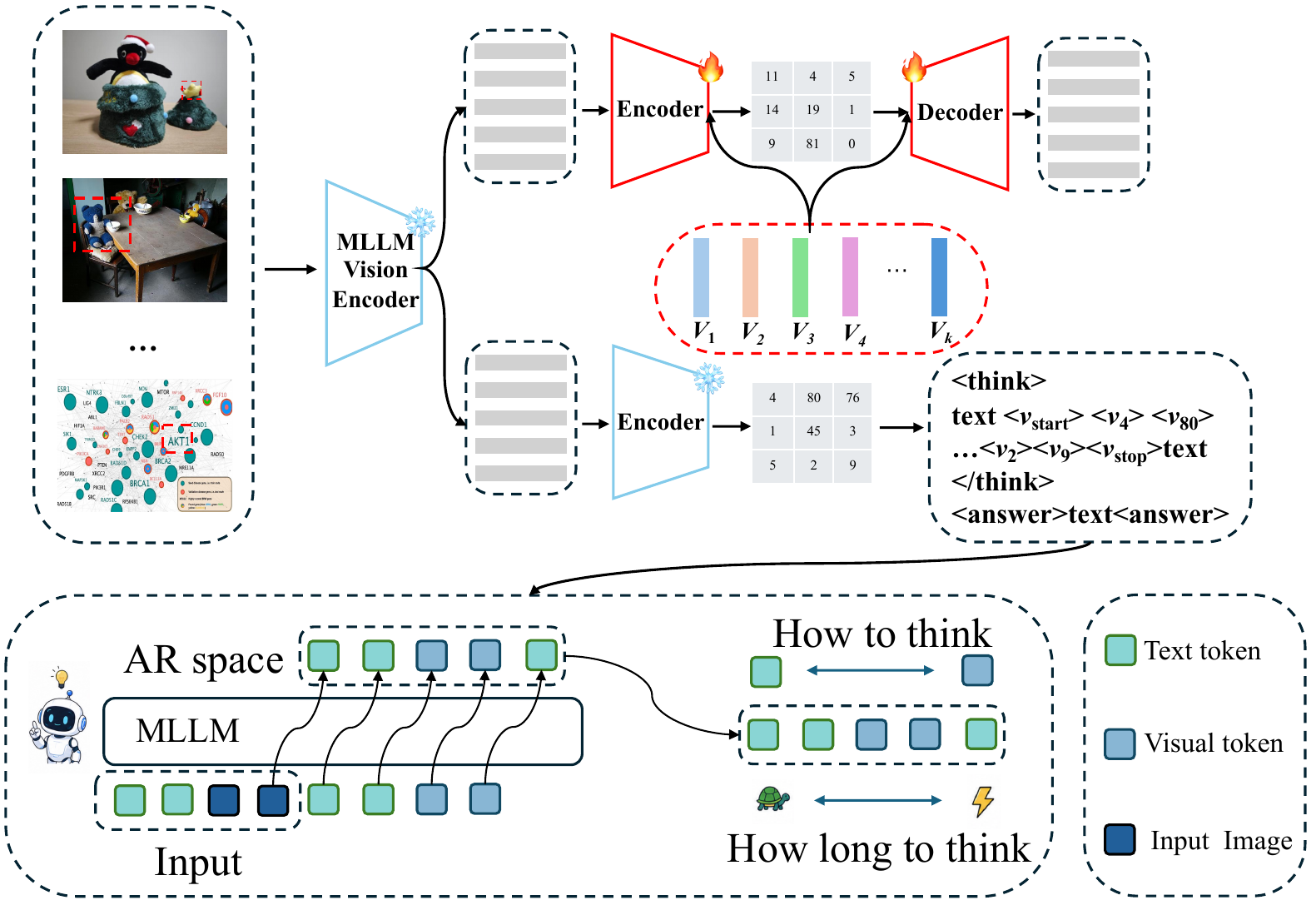}
    \caption{
        %VQ-VT converts tool-produced intermediate visual states into discrete visual-token sequences.
        VQ-VT converts tool-produced intermediate visual states into discrete visual-token sequences, enabling visual and textual reasoning to share a unified autoregressive token space for adaptive reasoning depth.
    }
    \label{fig:vqvae}
\end{figure}

As shown in Figure~\ref{fig:vqvae}, we introduce new latent visual tokens to facilitate visual-centric reasoning. To verify that these tokens encode meaningful visual information, we conduct a quantitative analysis based on image perturbation and random image replacement. Specifically, we compare the overlap between visual tokens generated from the original image and those generated from its perturbed version, using tokens generated from a randomly replaced image as a reference. If the original and perturbed images yield a higher token overlap than the original and randomly replaced images, this indicates that our visual tokens capture image-specific information and preserve relevant visual representations under perturbations, rather than merely reflecting generic pause token patterns\citep{goyal2024think}.
We propose a small experiment on 24 examples sampled from HRBench4K using the model. For each example, we kept the question and prompt fixed and changed only the image. The image conditions included a 2$\times$ upscale, horizontal flip, stretch, mild noise with a small mask, and the paired HRBench8K image, which may have a different crop. The unrelated image is the baseline. We also repeated the original image as a stability check.

We compared the generated latent visual token with normalized Levenshtein similarity,
\begin{equation}
  S(x,y)=1-\frac{d_{\mathrm{Lev}}(x,y)}{\max(|x|,|y|)},
  \label{eq:vq-edit-similarity-en}
\end{equation}
which supports sequences of different lengths of visual token. Higher values mean more similar sequences.
\begin{table}[t]
\centering
%\small
\caption{VQ code similarity to the original 4K image.}
\label{tab:vq-token-visual-similarity-en}
\resizebox{0.6\textwidth}{!}{
\begin{tabular}{lc}
\toprule
Image condition &  Similarity  \\
\midrule
Repeated original & 90.2\%  \\
Same image, 2$\times$ upscale  & 84.0\%  \\
stretch  & 83.0\%  \\
Horizontal flip  & 72.8\% \\
Noise + small mask  & 69.6\%  \\
Paired HRBench8K image  & 67.6\%  \\
Unrelated image  & 41.5\% \\
\bottomrule
\end{tabular}}
\end{table}

As shown in Table~\ref{tab:vq-token-visual-similarity-en}, most image changes that preserve the scene produced more similar visual token sequences than replacing the image. Even when replacing the input with a different image from the same question in HRBench 8K, the token overlap remains significantly higher than with a random image. This demonstrates that the latent visual tokens are sensitive to the input image itself and that the model indeed leverages visual information during reasoning.
\subsection{Analysis of Difficulty-aware Fast-Slow Thinking}
\label{sec:fast}
\paragraph{Theoretical Analysis.}
In the main paper, we restrict the comparison and length penalty to samples that receive the maximum reward, our framework is inherently more flexible. Specifically, it allows different reward thresholds to be specified, enabling even incorrectly routed samples to participate in the comparison while ensuring that samples with incorrect final answers never receive higher rewards than correct ones. This flexibility makes the proposed framework readily extensible to other domains with more complex reward functions.

Let
\begin{equation}
d_{ij}
=
\tanh\!\left(
\frac{L_i-L_j}
{\max(\sigma_L,L_{\min})+\epsilon}
\right).
\end{equation}
Since the denominator is positive and $\tanh(x)\in(-1,1)$,
we have $|d_{ij}|<1$. Thus, regardless of the length difference,
the magnitude of the compression term is bounded by $\beta$.

Let
\begin{equation}
\mathcal C_G=\{i:r_i>\tau_{\rm corr}\},
\qquad
\mathcal F_G=G\setminus\mathcal C_G
\end{equation}
denote the correct and incorrect subsets of group $G$. For any
$i\in\mathcal C_G$ and $k\in\mathcal F_G$, we have $r_i>r_k$.
Moreover, $c_{ik}=0$ and $h_{ik}=1$. Therefore, under the current
objective,
\begin{equation}
\Phi_{ik}
=
\begin{cases}
r_i-r_k, & \text{easy or medium group},\\
r_i-r_k+\beta_{\rm rep}[d_{ik}]_+,
& \text{hard group},
\end{cases}
\end{equation}
which is strictly positive. Hence, at the pairwise level, neither
compression nor repair can reverse the ordering between a correct
and an incorrect rollout.

We next consider the aggregated advantage
\begin{equation}
\bar A_i
=
\frac{1}{M-1}
\sum_{\substack{j\in G\\j\neq i}}\Phi_{ij},
\qquad M=|G|.
\end{equation}
Let $K=|\mathcal C_G|$ and define the within-group reward margin
\begin{equation}
\gamma_G
=
\min_{i\in\mathcal C_G}r_i
-
\max_{k\in\mathcal F_G}r_k.
\end{equation}
For an easy group, only correct--correct pairs receive the length
penalty. Consequently, for $i\in\mathcal C_G$ and
$k\in\mathcal F_G$,
\begin{align}
\bar A_i-\bar A_k
&=
\frac{M(r_i-r_k)}{M-1}
-
\frac{\beta}{M-1}
\sum_{\ell\in\mathcal C_G\setminus\{i\}}d_{i\ell}\\
&\ge
\frac{M\gamma_G-\beta(K-1)}{M-1}.
\end{align}
Therefore, a sufficient condition for preserving the ordering after
group aggregation is
\begin{equation}
\beta(K-1)<M\gamma_G.
\end{equation}
Since $K-1<M$, the simpler condition $\beta<\gamma_G$ is sufficient
for every possible group composition.

In the normalized binary-reward case, where correct rollouts receive
$r_i=1$ and incorrect rollouts satisfy
$r_k\le\tau_{\rm corr}$, we have
$\gamma_G\ge1-\tau_{\rm corr}$. It follows that
\begin{equation}
\beta<1-\tau_{\rm corr}
\end{equation}
is a group-independent sufficient condition ensuring that every
correct rollout receives a larger aggregated advantage than every
incorrect rollout. Medium groups contain no length term, while the
repair term in hard groups is nonnegative, so their correctness
ordering is preserved without this additional restriction.

Finally, the normalized advantage is
\begin{equation}
A_i=\frac{\bar A_i-\mu_G}{\sigma_G+\epsilon}.
\end{equation}
Because $\sigma_G+\epsilon>0$, this is a strictly increasing affine
transformation and therefore preserves the ordering of $\bar A_i$.

More generally, for continuous or RLHF-style rewards, let
$y_i\in\{0,1\}$ denote the correctness label and define
\begin{equation}
\gamma
=
\inf_{i:\,y_i=1}r_i
-
\sup_{j:\,y_j=0}r_j.
\end{equation}
The condition $\gamma>0$ means that the primary reward maintains a
strict separation between correct and incorrect rollouts.

Consider the generalized pairwise potential
\begin{equation}
\Phi_{ij}^{\mathrm{gen}}
=
\Delta r_{ij}
-
\beta_L g_{ij}^{L}
+
\sum_{m=1}^{P}\lambda_m g_{ij}^{(m)},
\end{equation}
where $\beta_L\geq0$, $|g_{ij}^{L}|\leq1$, and
$|g_{ij}^{(m)}|\leq1$. The auxiliary terms may represent reasoning
length, route consistency, response quality, style preference, or
normalized RLHF signals. Define their total worst-case influence as
\begin{equation}
B_{\mathrm{aux}}
=
\beta_L+\sum_{m=1}^{P}|\lambda_m|.
\end{equation}
For every correct--incorrect pair $(i,j)$, we then have
\begin{equation}
\Phi_{ij}^{\mathrm{gen}}
\geq
\gamma-B_{\mathrm{aux}}.
\end{equation}
Therefore, the pairwise correctness ordering is preserved whenever
\begin{equation}
B_{\mathrm{aux}}<\gamma.
\end{equation}

In the normalized unit-correctness-reward case, where correct
rollouts receive reward $1$ and incorrect rollouts receive at most
$\tau_{\rm corr}$, we have
\begin{equation}
\gamma\geq1-\tau_{\rm corr}.
\end{equation}
Consequently, the group-independent condition
\begin{equation}
\beta_L+\sum_{m=1}^{P}|\lambda_m|
<
1-\tau_{\rm corr}
\end{equation}
is sufficient to preserve pairwise correctness ordering.

This result interprets the correctness margin as an auxiliary reward
budget. The relative magnitudes of $\beta_L$ and $\lambda_m$ determine
the importance of length efficiency and other preferences, whereas
their total worst-case influence remains below the correctness margin.
Thus, auxiliary objectives may rank rollouts within the same
correctness level without making an incorrect rollout preferable to
a correct one.

For arbitrary nonseparable pairwise shaping terms, a conservative
group-independent condition for preserving the aggregated ordering is
$B_{\mathrm{aux}}<\gamma/2$.
\paragraph{Ablation on Thresholds}
We further evaluate the importance of difficulty discrimination based on the pass-rate threshold. In this ablation, we remove the threshold and instead adopt a strategy similar to previous work, where the shortest correct response is treated as the optimization target~\citep{pmlr-v267-yang25ae,yang2025thinkneedselfadaptivechainofthought,Wang_2026_CVPR}. As shown in Table~\ref{tab:thresholds}, removing the threshold leads to a degradation in performance. This finding indicates that distinguishing tasks of different difficulty is beneficial. In the next section, we further investigate the impact of slow thinking on tasks with varying levels of difficulty.
\begin{table}[htbp]
  \centering
  %\small
  \setlength{\tabcolsep}{4pt}
  \renewcommand{\arraystretch}{0.75}
   \caption{Ablation on thresholds.}
\resizebox{0.7\textwidth}{!}{
  \begin{tabular}{lccc}
  \toprule
  Components & Visual-Centric & Reasoning-Centric  & Total $\Delta$ \\
   \midrule
  w/o threshold & 70.5  & 56.0 &  -1.2 \\
  standard & 71.2  & 56.5 & -- \\
  \bottomrule
  \end{tabular}}
  
  \label{tab:thresholds}
  \end{table}

\subsection{Effectiveness of Long CoT}
\label{sec:long-cot-effectiveness}
\begin{figure*}[htbp]
    \centering

    \begin{subfigure}[t]{0.42\textwidth}
        \centering
        \includegraphics[
            width=\linewidth,
            height=3.4cm,
            keepaspectratio
        ]{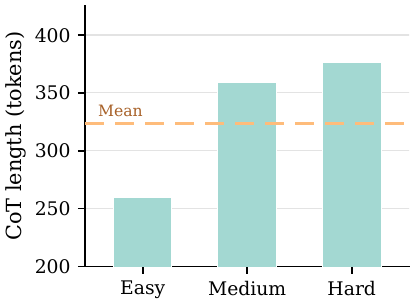}
        \caption{The average response length in different complexity.} % 建议修改你的标题
        \label{fig:difficult} 
    \end{subfigure}
    \hfill
    \begin{subfigure}[t]{0.57\textwidth}
        \centering
        \includegraphics[
            width=\linewidth,
            height=3.4cm,
            keepaspectratio
        ]{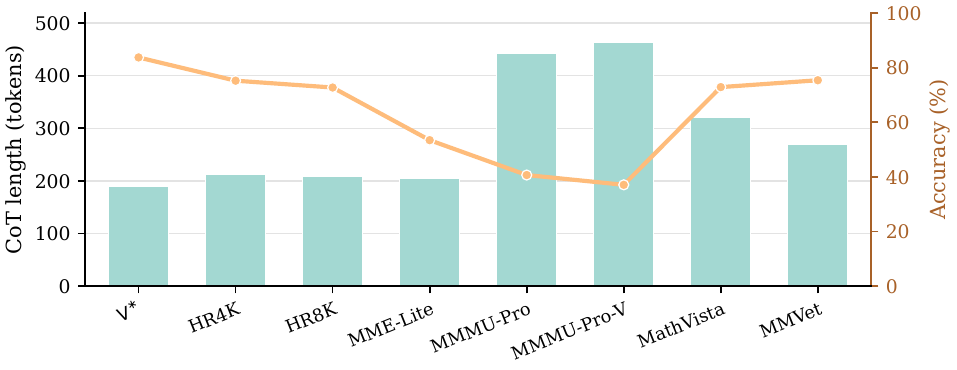}
    \caption{CoT length and accuracy across perception- and reasoning-oriented datasets.}
    \label{fig:datasetcot}
    \end{subfigure}

    \caption{Difficulty-dependent reasoning length and benchmark-level CoT statistics.}
    \label{fig:cot_length_benchmark}
\end{figure*}
\subsubsection{Effectiveness of Fast-slow thinking.}To investigate whether our model learns difficulty-aware reasoning, we analyze CoT length distributions under different difficulty levels. Samples are categorized by the correctness of our model and Qwen2.5-VL: both correct as easy, only one correct as medium, and both incorrect as hard. As shown in Figure~\ref{fig:difficult}, our model adaptively allocates reasoning budgets, generating the shortest CoTs for easy samples and the longest for hard samples. Medium samples receive moderate additional reasoning, suggesting that extra computation is beneficial when deeper reasoning is required. These results demonstrate that L2T learns adaptive reasoning allocation rather than simply favoring shorter or longer CoTs.
\subsubsection{CoT Length Across Benchmark}
We further analyze the relationship between CoT length and model performance across different benchmarks. As shown in Figure.~\ref{fig:datasetcot}, CoT length exhibits different patterns between perception- and reasoning-oriented tasks. On perception-centric benchmarks, we observe no clear correlation between reasoning length and accuracy, suggesting that increasing textual reasoning alone provides limited benefits when failures primarily originate from visual perception errors. In contrast, on reasoning-centric benchmarks, incorrect samples generally produce longer CoTs, indicating that challenging reasoning problems naturally require more extensive exploration and self-reflection. These observations highlight that different task types benefit from different forms of computation: visual tasks require effective perceptual computation, while reasoning tasks can benefit more from additional reasoning depth.
\subsubsection{Ablation on Collapsed CoT}To further investigate the role of CoT reasoning, we increase the number of Stage 3 training steps and remove the online filtering strategy, allowing a larger proportion of easy samples to dominate optimization and encouraging progressively shorter reasoning traces. As shown in Table~\ref{tab:overthinking}, under this setting, the CoT eventually collapses into a nearly fixed template of only about 10 tokens. Compared with the standard model, performance on perception-oriented benchmarks decreases by only 2.6\%, whereas reasoning-oriented benchmarks suffer a much larger drop of 12.2\%. These results suggest that long CoT provides limited benefit for perception tasks but remains crucial for reasoning-intensive problems.

  \begin{table}[htbp]
  \centering
  \small
  \setlength{\tabcolsep}{4pt}
  \renewcommand{\arraystretch}{0.75}
   \caption{Ablation on Route Reward}
   \resizebox{0.75\textwidth}{!}{
  \begin{tabular}{lccc}
  \toprule
  Category & standard (\%) & overtraining (\%) & $\Delta$ \\
  \midrule
  \multicolumn{4}{c}{Average Length of CoT} \\
  \midrule
  Visual-Centric & 206.8  & 10 & +196.8 \\
  Reasoning-Centric & 416.2  & 10 & +406.2  \\
  \midrule
  \multicolumn{4}{c}{Accuracy} \\
  \midrule
  Visual-Centric & 71.2  & 68.6 & +2.6 \\
  Reasoning-Centric & 56
  .5  & 44.3 & +12.2  \\
  \bottomrule
  \end{tabular}}

  \label{tab:overthinking}
  \end{table}

\subsubsection{CoT Length Across Failure Modes}

\begin{table}[t]
\centering
%\small
\setlength{\tabcolsep}{4pt}
\caption{Mean and median CoT length for each answer outcome and failure mode.}

\begin{tabular}{lrrr}
\toprule
Outcome / failure mode & $n$ & Mean & Median \\
\midrule
Correct-all & 4,573 & 277.3 & 214 \\
Correct-Perception & 2,370 & 200.6 & 198 \\
Correct-Reasoning & 2,203 & 359.9 & 260 \\
Perception failure & 1,758 & 241.9 & 216 \\
Knowledge failure & 517 & 430.0 & 281 \\
Reasoning-process failure & 724 & 505.8 & 304.5 \\
Mixed failure & 302 & 461.9 & 308 \\
Correct CoT, wrong output & 262 & 356.5 & 244 \\
Termination/completeness failure & 94 & 1,644.6 & 2,046 \\
\bottomrule
\end{tabular}

\label{tab:long-cot-failure-modes}
\end{table}
We further analyze the failure cases by categorizing the evaluation samples into correct predictions and three error types: perception failure, knowledge deficiency, and reasoning failure. We use GPT-5.5 to categorize the types of errors\citep{gpt5}.
Table~\ref{tab:long-cot-failure-modes} summarizes 8,230 responses. Interestingly, both the median and mean reasoning lengths of perception failures and knowledge-deficiency cases are comparable to those of correctly solved samples. Moreover, these failures typically emerge at the early stages of reasoning, rather than after extended deliberation. In contrast, reasoning failures are associated with substantially longer CoT traces and usually occur at later stages of the reasoning process. We conjecture that perception failures and knowledge deficiencies primarily stem from limitations in visual perception or internal knowledge, and therefore cannot be effectively remedied by simply extending the reasoning process. This observation also indicates that using rollout pass rate alone as a proxy for task difficulty is insufficient for adaptive CoT allocation, as it may unnecessarily encourage longer reasoning on tasks that are unlikely to benefit from additional computation. We leave a more fine-grained difficulty estimation strategy for future work.
  \begin{figure*}[htbp] % 如果是双栏跨栏
    \centering
    % 左侧图 1
    \begin{subfigure}[b]{0.49\textwidth}
        \centering
        % 【关键修改】：按绝对高度缩放，而不是宽度
        \includegraphics[width=1\textwidth]{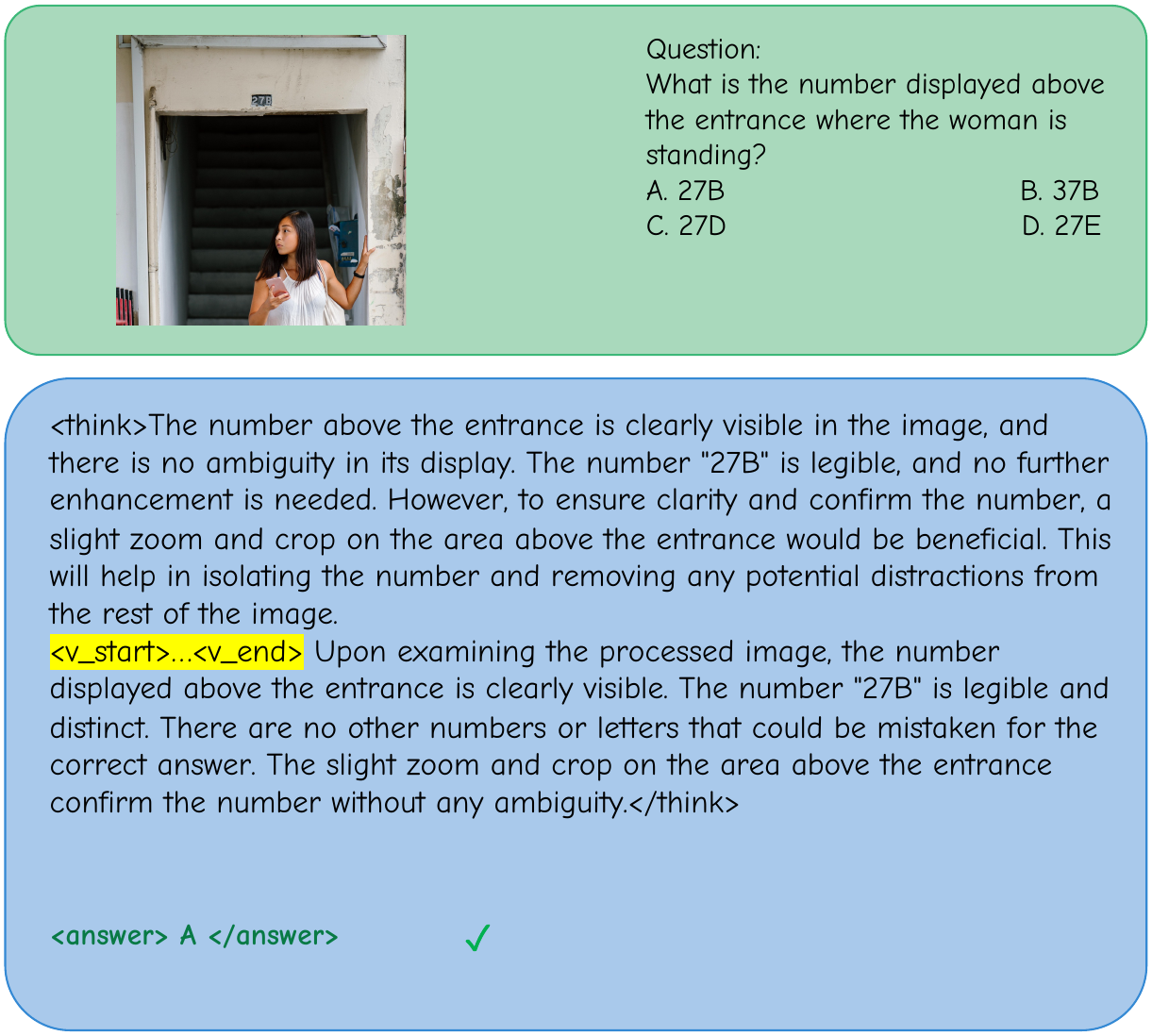} 
        \caption{Case study on visual sample.}
        \label{fig:case1}
    \end{subfigure}
    \hfill
    % 右侧图 2
    \begin{subfigure}[b]{0.49\textwidth}
        \centering
        % 【关键修改】：保持和左边完全一样的 height 数值
        \includegraphics[width=1\textwidth]{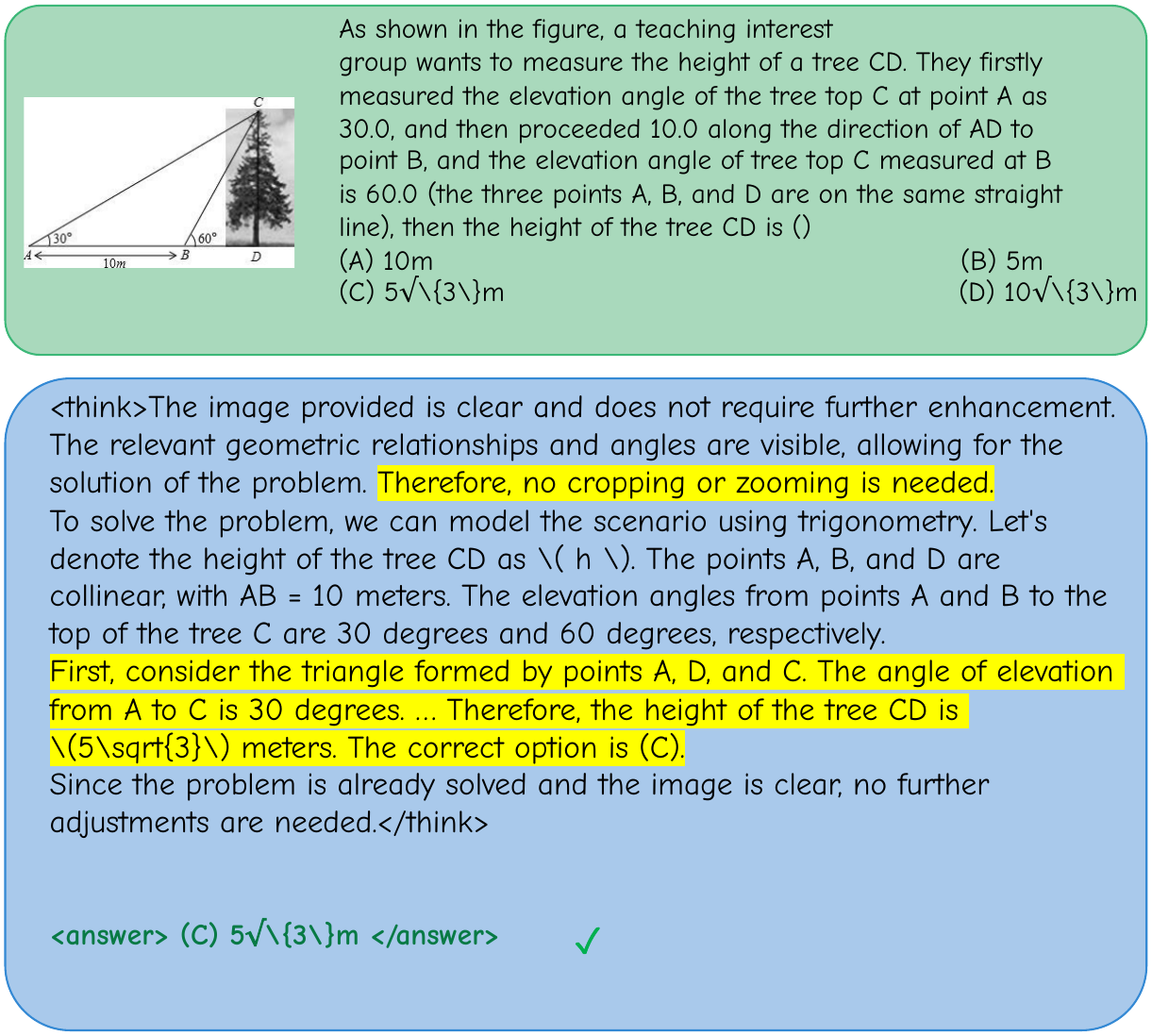} 
        \caption{Case study on reasoning sample.}
        \label{fig:case2}
    \end{subfigure}
    
    \caption{Successful case study of our method.}
    \label{fig:case_study_success}
\end{figure*}
\begin{figure*}[htbp] % 如果是双栏跨栏
    \centering
    % 左侧图 1
    \begin{subfigure}[b]{0.49\textwidth}
        \centering
        % 【关键修改】：按绝对高度缩放，而不是宽度
        \includegraphics[width=1\textwidth]{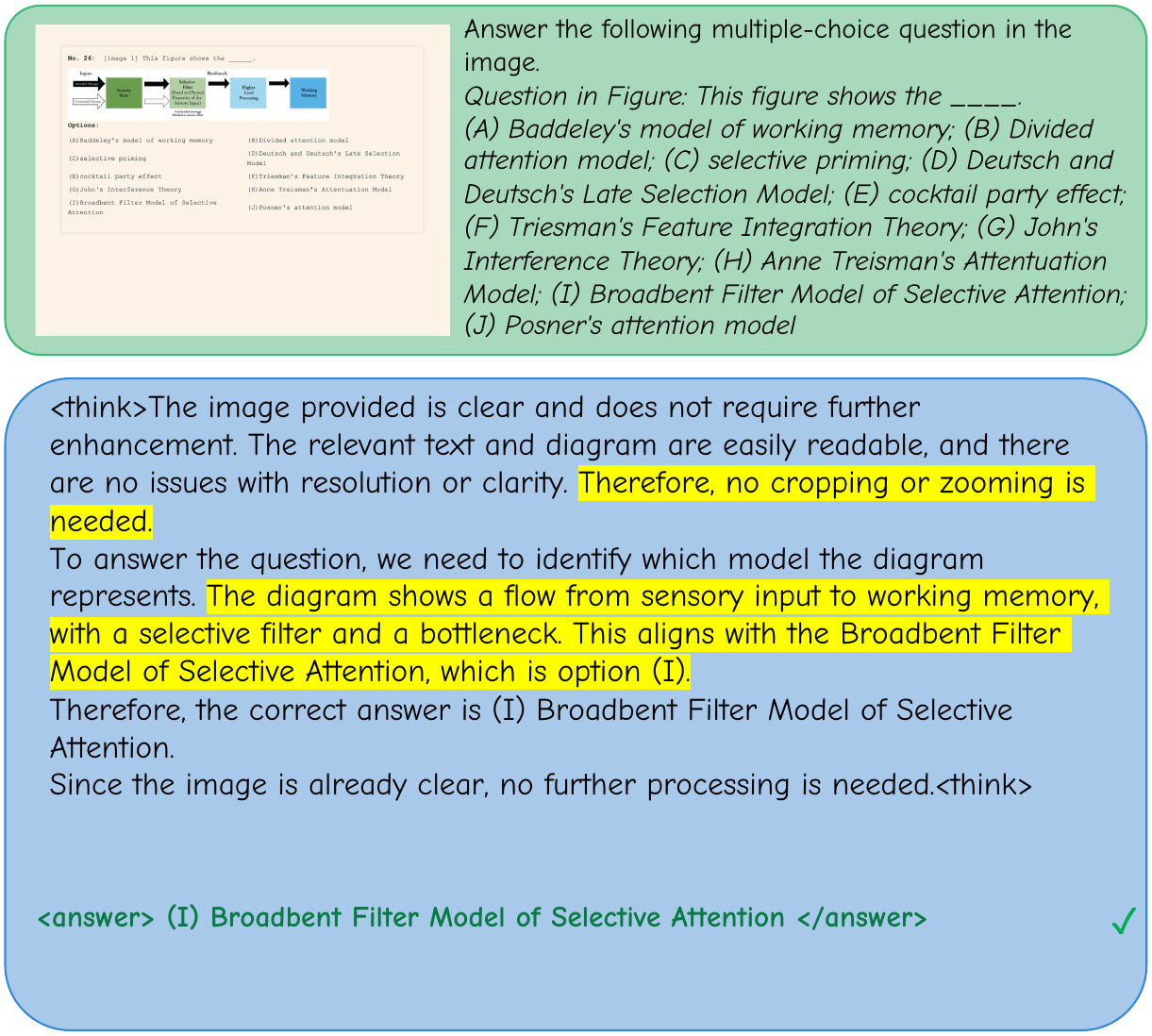} 
        \caption{Case study on interleaved sample.}
        \label{fig:case3}
    \end{subfigure}
    \hfill
    % 右侧图 2
    \begin{subfigure}[b]{0.49\textwidth}
        \centering
        % 【关键修改】：保持和左边完全一样的 height 数值
        \includegraphics[width=1\textwidth]{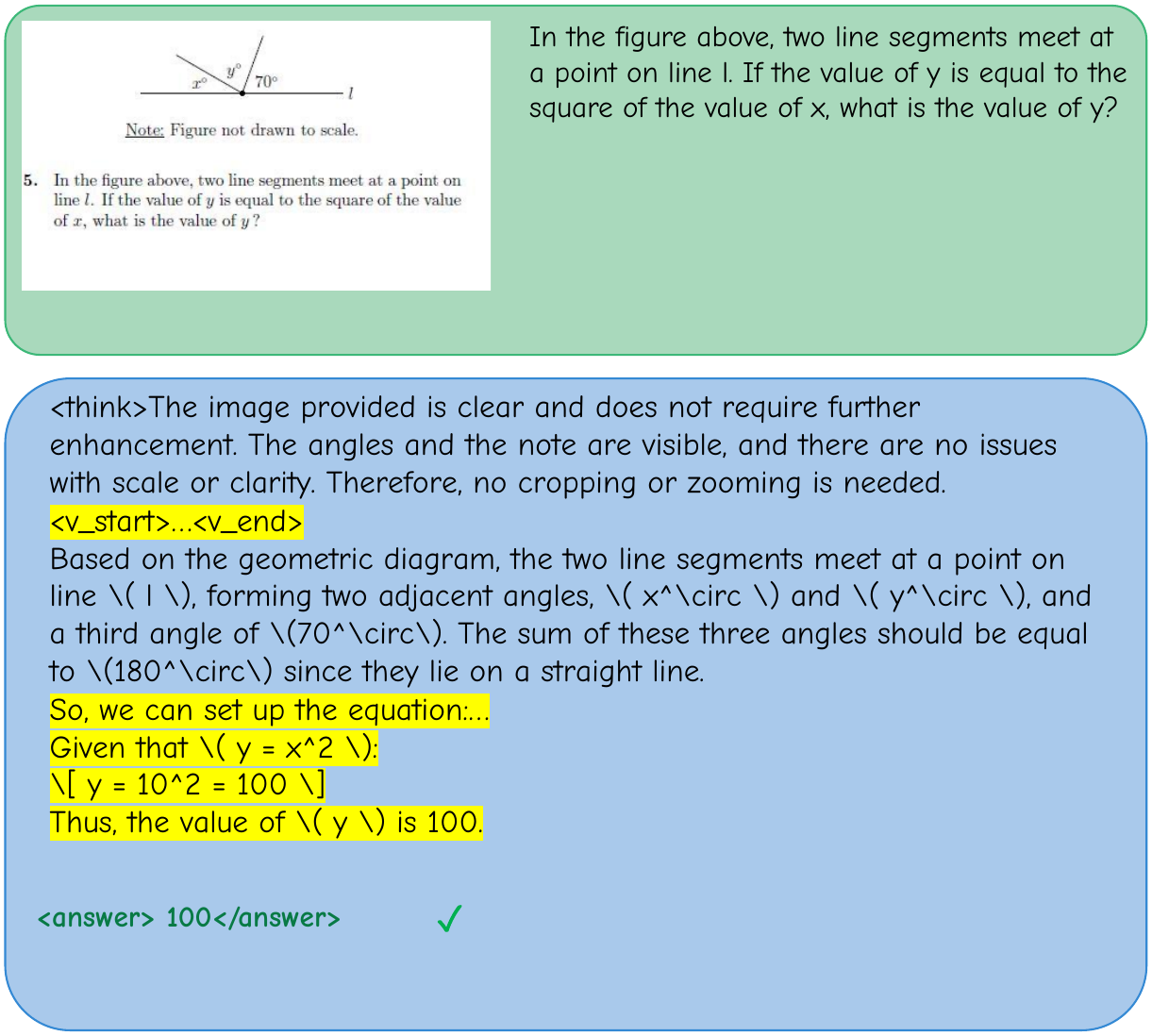} 
        \caption{Case study on knowledge-based sample.}
        \label{fig:case4}
    \end{subfigure}
    
    \caption{Successful case study of our method.}
    \label{fig:case_study_mixed}
\end{figure*}
\begin{figure*}[htbp] % 如果是双栏跨栏
    \centering
    % 左侧图 1
    \begin{subfigure}[b]{0.49\textwidth}
        \centering
        % 【关键修改】：按绝对高度缩放，而不是宽度
        \includegraphics[width=1\textwidth]{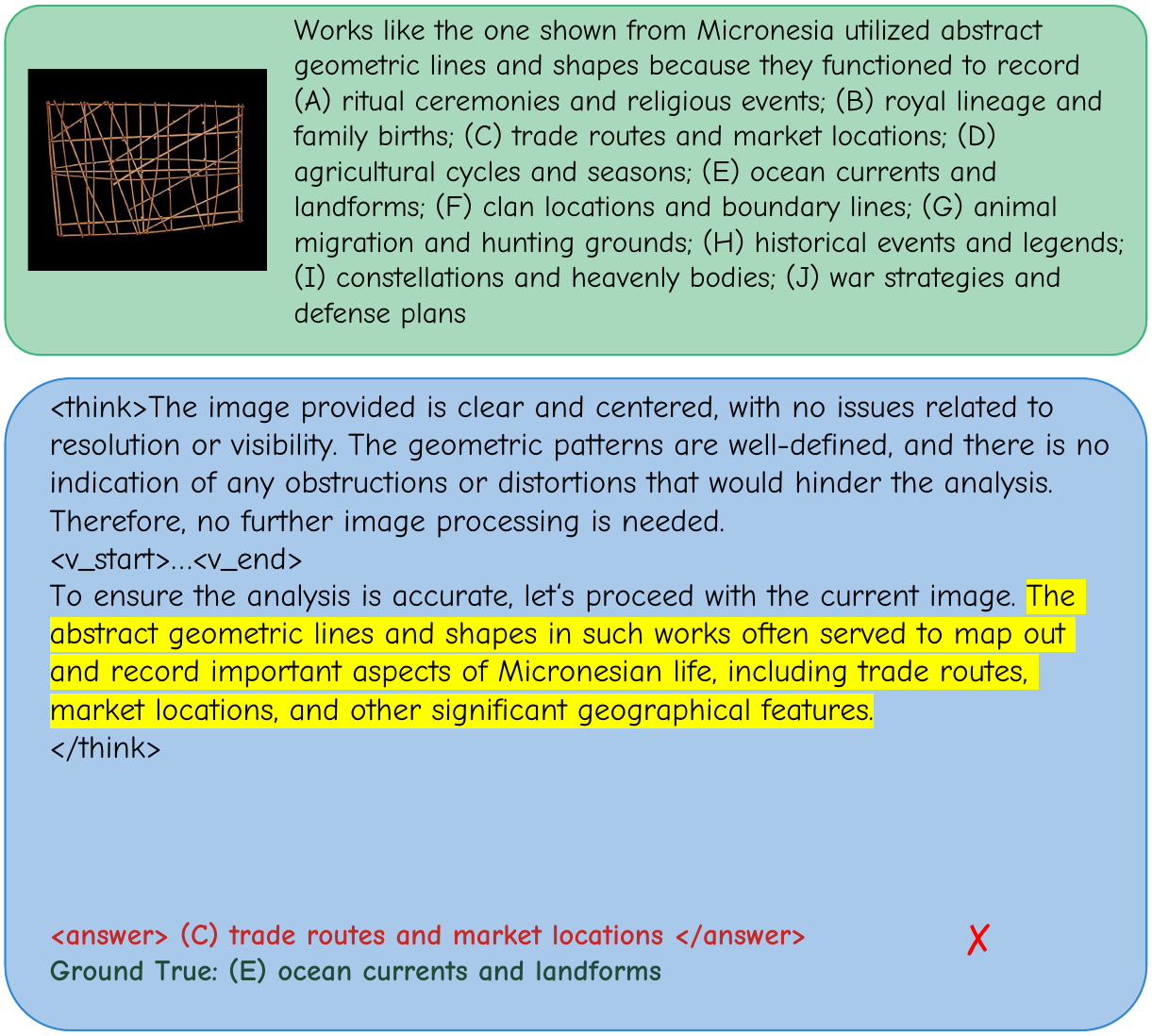} 
        \caption{Case study on knowledge-based sample.}
        \label{fig:case5}
    \end{subfigure}
    \hfill
    % 右侧图 2
    \begin{subfigure}[b]{0.49\textwidth}
        \centering
        % 【关键修改】：保持和左边完全一样的 height 数值
        \includegraphics[width=1\textwidth]{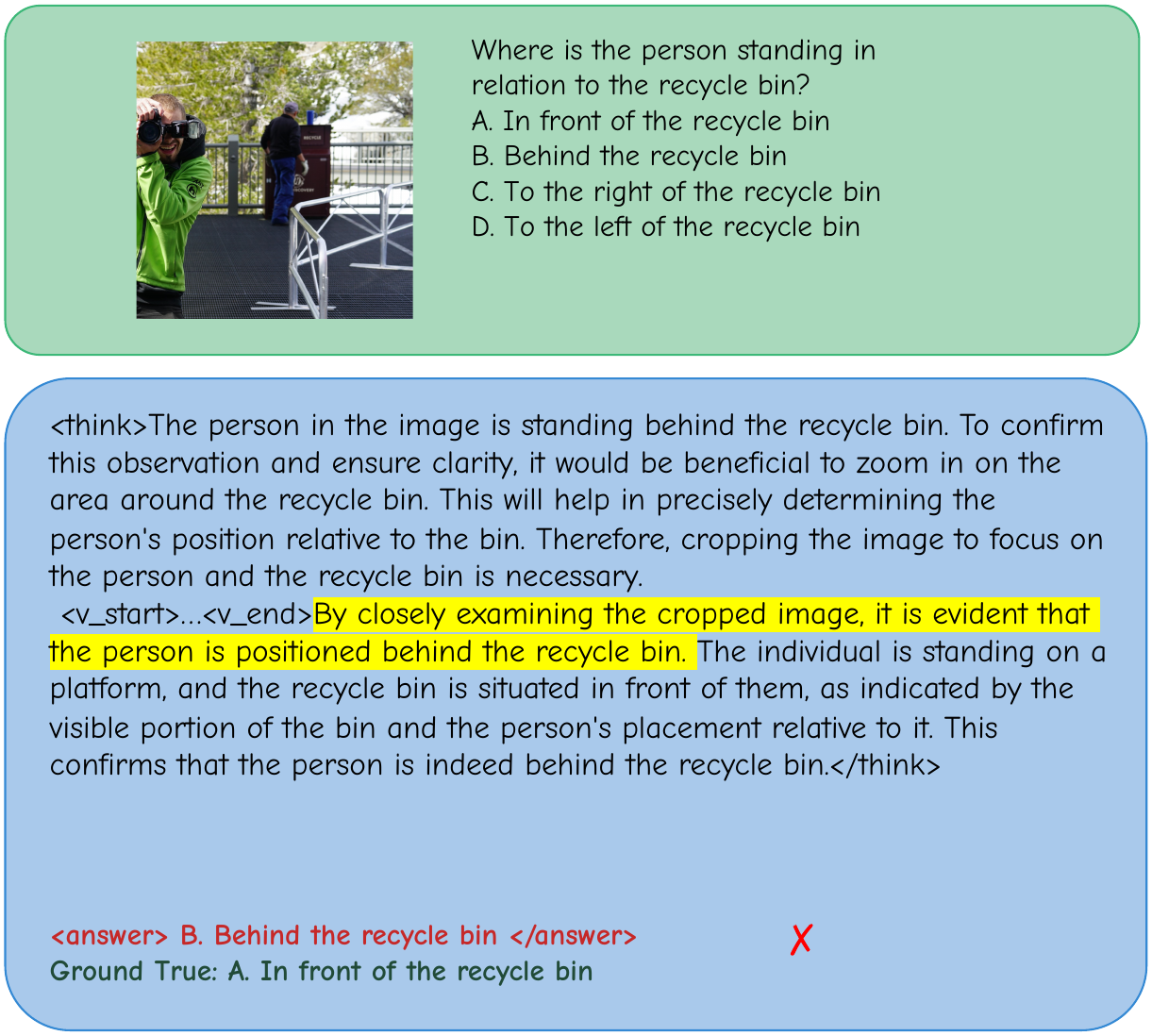} 
        \caption{Case study on visual sample.}
        \label{fig:case6}
    \end{subfigure}
    
    \caption{Failed case study of our method.}
    \label{fig:case_study_failure}
\end{figure*}
\begin{figure*}[htbp] % 如果是双栏跨栏
    \centering
    % 左侧图 1
    \begin{subfigure}[b]{0.49\textwidth}
        \centering
        % 【关键修改】：按绝对高度缩放，而不是宽度
        \includegraphics[width=1\textwidth]{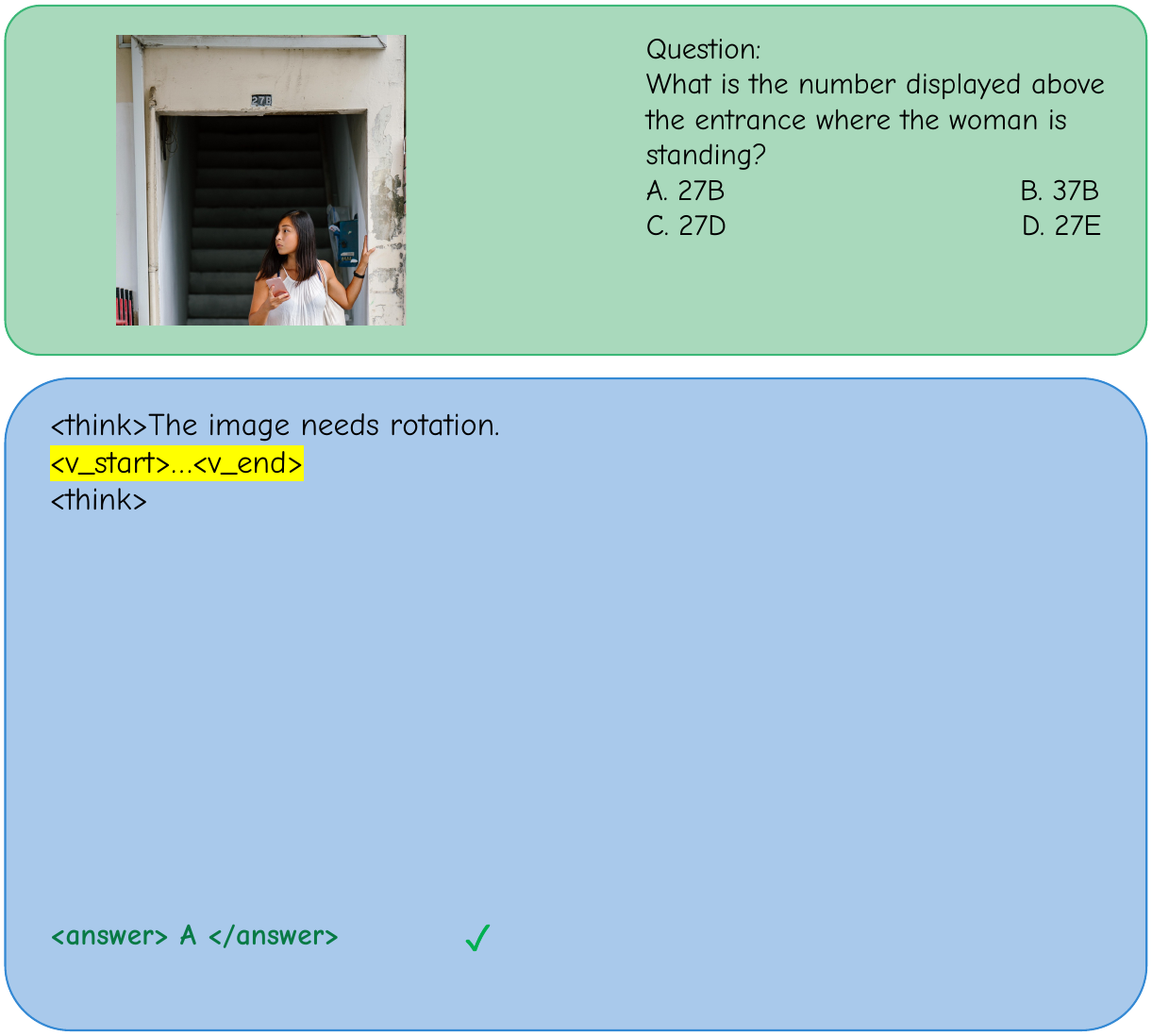} 
        \caption{Case study on visual sample.}
        \label{fig:case7}
    \end{subfigure}
    \hfill
    % 右侧图 2
    \begin{subfigure}[b]{0.49\textwidth}
        \centering
        % 【关键修改】：保持和左边完全一样的 height 数值
        \includegraphics[width=1\textwidth]{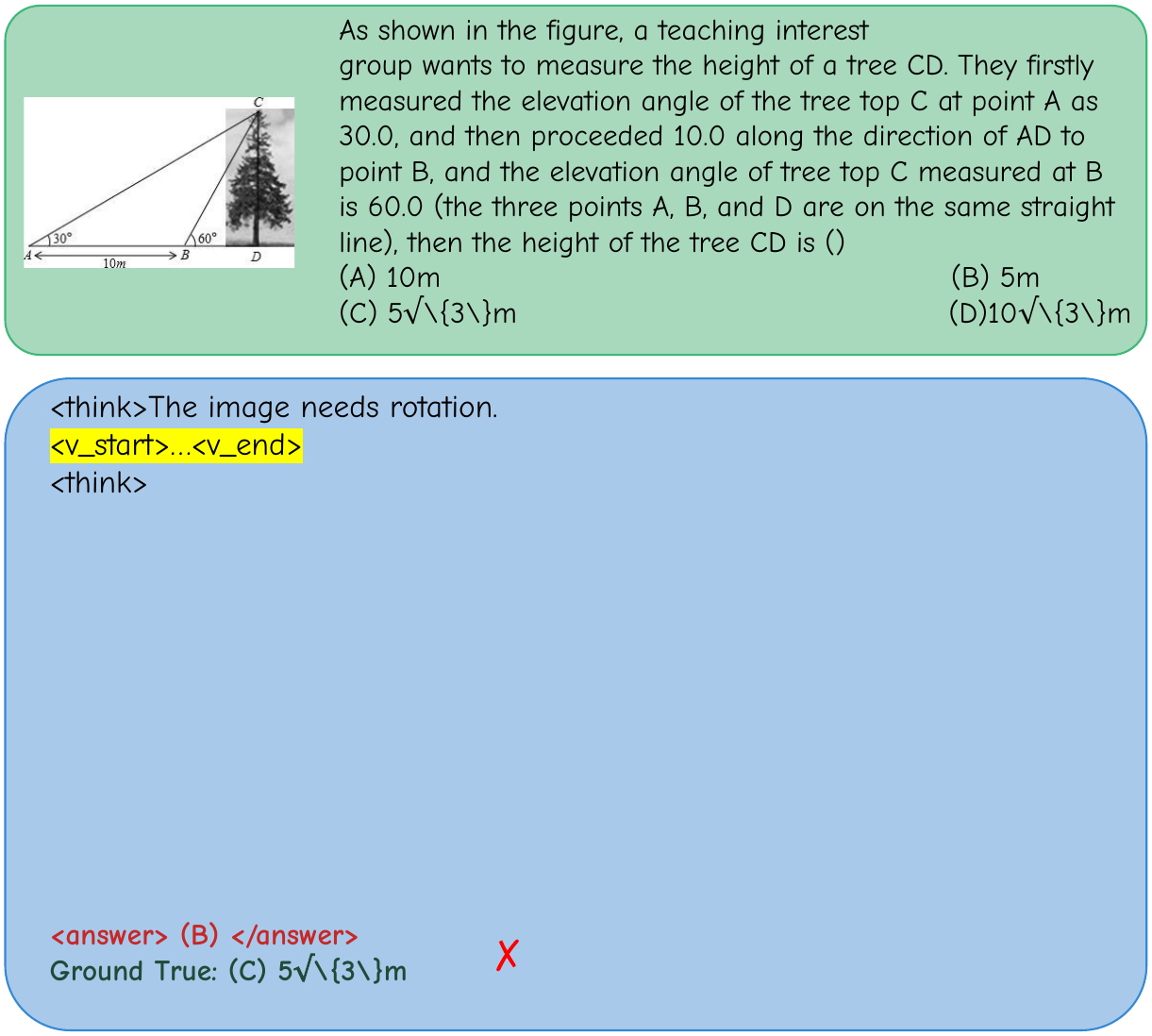} 
        \caption{Case study on reasoning sample.}
        \label{fig:case8}
    \end{subfigure}
    
    \caption{Case study on Collapsed CoT.}
    \label{fig:case_study_cot_collapse}
\end{figure*}

\subsection{Case Study}
\paragraph{Successful Case}
In this section, we present representative successful cases. Figure~\ref{fig:case1} illustrates a perception-oriented task in which the model correctly identifies the task type and generates appropriate visual tokens to facilitate visual perception. Figure~\ref{fig:case2} shows a reasoning-oriented example that is solved solely through textual reasoning without invoking visual tokens. Figure~\ref{fig:case3} presents a knowledge-based question, where the model answers directly from its internal knowledge without generating either visual tokens or unnecessary textual reasoning. Figure~\ref{fig:case4} demonstrates a task requiring both perception and reasoning, where the model first employs visual tokens to enhance visual representations and subsequently performs textual reasoning to derive the final answer.

\paragraph{Failed Case}
In this section, we analyze representative failure cases. In Figure~\ref{fig:case5}, the question primarily depends on the model's internal knowledge and therefore does not require additional reasoning. However, the model unnecessarily generates visual tokens and ultimately produces an incorrect answer due to insufficient internal knowledge. In Figure~\ref{fig:case6}, the model makes an incorrect perception at the beginning of the reasoning process. Although it subsequently generates visual tokens to re-examine the image, it fails to correct the initial perceptual error.

These cases suggest that extending the reasoning process provides limited benefit for correcting knowledge deficiencies or perceptual errors, consistent with our quantitative observations that such failure cases exhibit reasoning lengths comparable to those of correctly solved samples. We attribute these failures primarily to intrinsic limitations of the underlying model, namely insufficient domain knowledge and inadequate visual perception capabilities. When the required prior knowledge is absent, textual reasoning alone cannot compensate for the missing information to produce correct answers. Likewise, when visual perception is fundamentally inadequate, reprocessing images through the thinking-with-image paradigm is unlikely to recover the missing visual evidence. Therefore, future improvements should focus on enriching the model's foundational knowledge and strengthening the visual perception capability of the vision encoder, rather than relying solely on more sophisticated reasoning strategies.
\paragraph{Case on Collapsed CoT}
In this section, we investigate the behavior of the model after overtraining, where the CoT collapses into a short, fixed reasoning template. We revisit the previous case studies for comparison. Figure~\ref{fig:case1} and Figure~\ref{fig:case7} present the model's responses to a perception-oriented task before and after CoT collapse, respectively. The model correctly answers the question in both cases. In contrast, Figure~\ref{fig:case2} and Figure~\ref{fig:case8} compare the responses to a reasoning-intensive task before and after CoT collapse. Once the reasoning process is lost, the model fails to produce the correct answer. These observations suggest that different task types exhibit different levels of dependence on long CoT reasoning. In particular, reasoning-intensive tasks are substantially more affected by CoT collapse than perception-oriented tasks.

\subsection{Pseudo Code}
\label{sec:pseudo}
In this section, we present the pseudocode of the proposed algorithm to provide a clearer and more intuitive overview of the overall training procedure.
\begin{algorithm}[htbp]
\caption{VQ-VT Training}
\label{alg:vqvt}
\small
\begin{algorithmic}[1]
\Require Images $\mathcal D_{\mathrm{img}}$; frozen $E_{\mathrm{vis}}$;
  trainable $E_{\omega}$ and $D_{\phi}$
\Require Codebook $\mathcal V=\{e_j\}_{j=1}^{K}$; coefficient $\beta$
\Ensure Quantizer $Q\!\circ E_{\omega}$ and $K$ visual tokens
\While{the VQ-VT objective has not converged}
  \State Sample a minibatch of images $I\sim\mathcal D_{\mathrm{img}}$
  \State $\{h_i\}_{i=1}^{N}\gets E_{\mathrm{vis}}(I)$
    \Comment{frozen patch features}
  \For{$i=1,\ldots,N$}
    \State $z_i\gets \mathrm{Normalize}_{\ell_2}(E_{\omega}(h_i))$
    \For{$j=1,\ldots,K$}
      \State $\bar e_j\gets \mathrm{Normalize}_{\ell_2}(e_j)$
    \EndFor
    \State $k_i\gets \arg\max_{1\le j\le K} z_i^{\top}\bar e_j$
    \State $q_i\gets \bar e_{k_i}$
      \Comment{Eq.~(1)}
    \State $q_i^{\mathrm{ST}}\gets z_i+\sg(q_i-z_i)$
      \Comment{STE}
  \EndFor
  \State $\ell_i^{\mathrm{rec}}\gets
    \lVert h_i-D_{\phi}(q_i^{\mathrm{ST}})\rVert_2^2$
  \State $\ell_i^{\mathrm{code}}\gets
    \lVert\sg(z_i)-q_i\rVert_2^2$
  \State $\ell_i^{\mathrm{com}}\gets
    \beta\lVert z_i-\sg(q_i)\rVert_2^2$
  \State $\ell_i\gets
    \ell_i^{\mathrm{rec}}+\ell_i^{\mathrm{code}}+\ell_i^{\mathrm{com}}$
  \State $\displaystyle \mathcal L_{\mathrm{VQ}}\gets
    \frac{1}{N}\sum_{i=1}^{N}\ell_i$
    \Comment{Eq.~(2)}
  \State Update $E_{\omega}$, $D_{\phi}$, and $\mathcal V$ by descending
    $\nabla\mathcal L_{\mathrm{VQ}}$
\EndWhile
\State Add $\{\vstart,\texttt{<v1>},\ldots,\texttt{<vK>},\vend\}$ to the
  MLLM vocabulary
\State \Return $E_{\omega},Q,D_{\phi}$ and the extended vocabulary
\end{algorithmic}
\end{algorithm}

\begin{algorithm}[htbp]
\caption{Tool-Trajectory Discretization}
\label{alg:data}
\small
\begin{algorithmic}[1]
\Require Tool data $\mathcal D_{\mathrm{tool}}$; VQ-VT encoder
  $E_{\omega}$; quantizer $Q$
\Ensure Discrete visual-reasoning dataset $\mathcal D_{\mathrm{VT}}$
\State $\mathcal D_{\mathrm{VT}}\gets\varnothing$
\ForAll{$(X,I,C,A;t_{\mathrm{pre}},t_{\mathrm{post}})
  \in\mathcal D_{\mathrm{tool}}$}
  \State $I'\gets\Call{Execute}{C,I}$
    \Comment{Eq.~(3)}
  \State $(v_1,\ldots,v_m)\gets Q(E_{\omega}(I'))$
    \Comment{Eq.~(4)}
  \State $\mathbf v\gets(\vstart,v_1,\ldots,v_m,\vend)$
  \State $d\gets(X,t_{\mathrm{pre}},\mathbf v,t_{\mathrm{post}},A)$
    \Comment{Eq.~(5)}
  \State Add $d$ to $\mathcal D_{\mathrm{VT}}$
\EndFor
\State \Return $\mathcal D_{\mathrm{VT}}$
\end{algorithmic}
\end{algorithm}

\begin{algorithm}[htbp]
\caption{Three-Stage Adaptive Training}
\label{alg:three-stage}
\small
\begin{algorithmic}[1]
\Require Policy $\pi_{\theta}$; visual data $\mathcal D_{\mathrm{VT}}$;
  RL data $\mathcal D_{\mathrm{RL}}$
\Require Group size $M$; route and length hyperparameters
\Ensure Adaptively trained policy $\pi_{\theta}$
\Statex \textbf{Stage 1: visual supervised fine-tuning.}
\While{the SFT objective has not converged}
  \State Sample a minibatch $(X,I,s)\sim\mathcal D_{\mathrm{VT}}$
  \State $\displaystyle
    \mathcal L_{\mathrm{SFT}}\gets-\sum_{k}\log
    \pi_{\theta}(s_k\mid s_{<k},X,I)$
    \Comment{Eq.~(6)}
  \State Update $\theta$ by descending $\nabla\mathcal L_{\mathrm{SFT}}$
\EndWhile

\Statex \textbf{Stage 2: stable-reward GRPO.}
\While{the stable reasoning policy has not converged}
  \State Sample $(x,a^*)\sim\mathcal D_{\mathrm{RL}}$ and draw
    $G=\{y_i\}_{i=1}^{M}\sim\pi_{\theta}(\cdot\mid x)$
  \ForAll{$y_i\in G$}
    \State $(z_i,\hat a_i)\gets\Call{ParseResponse}{y_i}$
    \State $r_{\mathrm{acc},i}\gets
      \ind[\Call{Verify}{\hat a_i,a^*}]$
    \State $r_{\mathrm{fmt},i}\gets\ind[y_i\in\mathcal F]$
    \State $r_{\mathrm{stable},i}\gets
      r_{\mathrm{acc},i}r_{\mathrm{fmt},i}$
      \Comment{Eqs.~(7)--(9)}
  \EndFor
  \State Update $\pi_{\theta}$ by standard GRPO using
    $\{r_{\mathrm{stable},i}\}_{i=1}^{M}$
\EndWhile

\Statex \textbf{Stage 3: routing + fast--slow GRPO.}
\While{the adaptive policy has not converged}
  \State Sample $(x,a^*,n_v,n_t)\sim\mathcal D_{\mathrm{RL}}$ and draw
    $G=\{y_i\}_{i=1}^{M}\sim\pi_{\theta}(\cdot\mid x)$
  \ForAll{$y_i\in G$}
    \State $(R_i,L_i)\gets\Call{RoutingReward}{y_i,a^*,n_v,n_t}$
      \Comment{Alg.~\ref{alg:route}}
  \EndFor
  \State $\mathbf A\gets
    \Call{FastSlowAdvantage}{\{R_i,L_i\}_{i=1}^{M}}$
      \Comment{Alg.~\ref{alg:fastslow}}
  \State $\mathcal L_{\mathrm{GRPO}}\gets
    \Call{GRPOObjective}{x,G,\mathbf A}$
    \Comment{Alg.~\ref{alg:grpo}}
  \State Update $\theta$ by descending $\nabla\mathcal L_{\mathrm{GRPO}}$
\EndWhile
\State \Return $\pi_{\theta}$
\end{algorithmic}
\end{algorithm}

\begin{algorithm}[htbp]
\caption{Modality-Aware Routing Reward}
\label{alg:route}
\small
\begin{algorithmic}[1]
\Function{RoutingReward}{$y,a^*,n_v,n_t$}
\Require Thresholds $\tau_v,\tau_t,\tau_v^{\mathrm{mix}},
  \tau_t^{\mathrm{mix}},V_{\min},T_{\min}$
\Require Penalties $\lambda_{\mathrm{novis}},
  \lambda_{\mathrm{simple}}$
  \State $(z,\hat a)\gets\Call{ParseResponse}{y}$
  \State $V\gets V(z)$; $T\gets T(z)$; $L\gets V+T$
  \State $\rho_v\gets V/(V+T)$;
    $\rho_t\gets T/(V+T)$
    \Comment{Eq.~(10)}
  \Statex $\triangleright$ A valid trace has $V+T>0$.
  \State $\mathcal S(\rho,N;\tau,N_0)\gets
    \min(\rho/\tau,1)\min(N/N_0,1)$
  \State $s_v\gets\mathcal S(\rho_v,V;\tau_v,V_{\min})$
    \Comment{Eq.~(11)}
  \State $s_t\gets\mathcal S(\rho_t,T;\tau_t,T_{\min})$
    \Comment{Eq.~(12)}
  \State $s_v^{\mathrm{mix}}\gets
    \mathcal S(\rho_v,V;\tau_v^{\mathrm{mix}},V_{\min})$
    \Comment{Eq.~(13)}
  \State $s_t^{\mathrm{mix}}\gets
    \mathcal S(\rho_t,T;\tau_t^{\mathrm{mix}},T_{\min})$
    \Comment{Eq.~(14)}
  \If{$n_v=1$ and $n_t=0$}
    \State $s_{\mathrm{route}}\gets s_v$
  \ElsIf{$n_v=0$ and $n_t=1$}
    \State $s_{\mathrm{route}}\gets
      s_t\lambda_{\mathrm{novis}}^{\ind[V>0]}$
  \ElsIf{$n_v=1$ and $n_t=1$}
    \State $s_{\mathrm{route}}\gets
      \min(s_v^{\mathrm{mix}},s_t^{\mathrm{mix}})$
  \Else
    \State $s_{\mathrm{route}}\gets
      \lambda_{\mathrm{simple}}^{\ind[V>0]}$
  \EndIf
  \Statex \hfill $\triangleright$ Four routing types in Eq.~(15)
  \State $r_{\mathrm{acc}}\gets\ind[\Call{Verify}{\hat a,a^*}]$
  \State $r_{\mathrm{fmt}}\gets\ind[y\in\mathcal F]$
  \State $R\gets(0.9r_{\mathrm{acc}}+0.1s_{\mathrm{route}})
    r_{\mathrm{fmt}}$
    \Comment{Eq.~(16)}
  \State \Return $(R,L)$
\EndFunction
\end{algorithmic}
\end{algorithm}

\begin{algorithm}[htbp]
\caption{Difficulty-Aware Fast--Slow Advantage}
\label{alg:fastslow}
\small
\begin{algorithmic}[1]
\Function{FastSlowAdvantage}{$\{R_i,L_i\}_{i=1}^{M}$}
\Require $\tau_{\mathrm{corr}}=0.99$;
  $\tau_{\mathrm{easy}},\tau_{\mathrm{hard}}$
\Require Length floor $L_{\min}$; coefficients $\beta,\beta'$
  \For{$i=1,\ldots,M$}
    \State $q_i\gets\ind[R_i>\tau_{\mathrm{corr}}]$
  \EndFor
  \State $p_G\gets M^{-1}\sum_{i=1}^{M}q_i$
    \Comment{Eq.~(17)}
  \State $\sigma_L\gets\Std(\{L_i\}_{i=1}^{M})$;
    $d_L\gets\max(\sigma_L,L_{\min})$
  \ForAll{ordered pairs $(i,j)$ with $i\ne j$}
    \State $\Delta R_{ij}\gets R_i-R_j$
    \State $\Delta L_{ij}\gets\tanh((L_i-L_j)/d_L)$
      \Comment{Eq.~(18)}
    \State $c_{ij}\gets q_iq_j$; $h_{ij}\gets q_i(1-q_j)$
      \Comment{Eqs.~(19)--(20)}
    \If{$p_G\ge\tau_{\mathrm{easy}}$}
      \State $\Phi_{ij}\gets
        \Delta R_{ij}-\beta\Delta L_{ij}c_{ij}$
        \Comment{easy}
    \ElsIf{$p_G\ge\tau_{\mathrm{hard}}$}
      \State $\Phi_{ij}\gets\Delta R_{ij}$
        \Comment{medium}
    \Else
      \State $\Phi_{ij}\gets
        \Delta R_{ij}+\beta'[\Delta L_{ij}]_+h_{ij}$
        \Comment{hard}
    \EndIf
    \Statex \hfill $\triangleright$ Pairwise potential in Eq.~(21)
  \EndFor
  \For{$i=1,\ldots,M$}
    \State $\displaystyle A_i\gets\Norm_G\!\left(
      \frac{1}{M-1}\sum_{j\ne i}\Phi_{ij}\right)$
      \Comment{Eq.~(22)}
  \EndFor
  \State \Return $\{A_i\}_{i=1}^{M}$
\EndFunction
\end{algorithmic}
\end{algorithm}
\begin{algorithm}[htbp]
\caption{Clipped GRPO Update}
\label{alg:grpo}
\small
\begin{algorithmic}[1]
\Function{GRPOObjective}{$x,G,\mathbf A$}
\Require Policies $\pi_\theta,\pi_{\theta_{\mathrm{old}}},
  \pi_{\mathrm{ref}}$
\Require Clip radius $\epsilon$; KL weight $\lambda_{\mathrm{KL}}$
  \ForAll{$y_{i,t}$ in rollout $y_i\in G$}
    \State $p_{i,t}\gets\pi_\theta(y_{i,t}\mid x,y_{i,<t})$
    \State $p^{\mathrm{old}}_{i,t}\gets
      \pi_{\theta_{\mathrm{old}}}(y_{i,t}\mid x,y_{i,<t})$
    \State $\rho_{i,t}\gets p_{i,t}/p^{\mathrm{old}}_{i,t}$
    \State $g^{\mathrm{raw}}_{i,t}\gets\rho_{i,t}A_i$
    \State $\bar\rho_{i,t}\gets
      \operatorname{clip}(\rho_{i,t},1-\epsilon,1+\epsilon)$
    \State $g^{\mathrm{clip}}_{i,t}\gets\bar\rho_{i,t}A_i$
    \State $g_{i,t}\gets
      \min(g^{\mathrm{raw}}_{i,t},g^{\mathrm{clip}}_{i,t})$
  \EndFor
  \For{$i=1,\ldots,M$}
    \State $\displaystyle \bar g_i\gets
      \frac{1}{|y_i|}\sum_t g_{i,t}$
  \EndFor
  \State $\displaystyle \mathcal L_{\mathrm{policy}}\gets
    -\frac{1}{M}\sum_{i=1}^{M}\bar g_i$
  \State $K\gets D_{\mathrm{KL}}(\pi_\theta\Vert\pi_{\mathrm{ref}})$
  \State $\mathcal L_{\mathrm{GRPO}}\gets
    \mathcal L_{\mathrm{policy}}+\lambda_{\mathrm{KL}}K$
    \Comment{Eq.~(23)}
  \State \Return $\mathcal L_{\mathrm{GRPO}}$
\EndFunction
\end{algorithmic}
\end{algorithm}
\newpage
\subsection{Experimental Details}
\label{detail}
\subsubsection{overall}
\paragraph{Training Data and Setup.}
Our experiments are based on Qwen2.5-VL~\citep{bai2025qwen25vltechnicalreport} .Our SFT data consists of Thyme-SFT~\citep{zhang2025thyme}, Zebra-CoT~\citep{li2025zebracotdatasetinterleavedvision} comes from Monet-SFT~\citep{wang2025monetreasoninglatentvisual}, and Vision-R1-SFT~\citep{huang2026visionr}.
For Thyme-SFT and Monet-SFT, the intermediate auxiliary reasoning images are converted into discrete visual tokens using VQ-VT, and the model is trained with LLaMA-Factory~\citep{zheng-etal-2024-llamafactory}. 
We further mix these data with the Mulberry dataset~\citep{yao2026mulberry} from Vision-R1-SFT~\citep{huang2026visionr}.
For reinforcement learning, we construct the RL dataset by combining Thyme-RL~\citep{zhang2025thyme}, ViRL39K~\citep{wang2026vlrethinker}, and Geo3K~\citep{lu-etal-2021-inter}, and train the model using EasyR1~\citep{zheng2025easyr1}.
During RL training, we enable online filtering to ensure a consistent rollout batch size throughout training, we optimize the policy using
GRPO~\citep{Guo_2025} in both Stage 2 and Stage 3. The SFT stage was conducted on 8 Nvidia H200 GPUs, while the RL and evaluation stages were conducted on 8 Nvidia H100 GPUs.

During the SFT stage, the models use 1,024 visual tokens. The vision tower and multimodal projector are frozen, while only the language model is trained. We adopt a conversational template consisting of system and user prompts, with the loss computed exclusively on the completion. The maximum sequence length is set to 8,196. The per-device training batch size is 8, with one gradient accumulation step, resulting in an effective global batch size of 64. We use the AdamW optimizer with a learning rate of 1×10$^{-5}$, a cosine learning rate scheduler, and two training epochs. Training is performed using BF16 precision and DeepSpeed ZeRO-3.

During the RL stage, we employ the GRPO algorithm with a training batch size of 256. Each prompt generates 16 candidate responses, and the PPO mini-batch size is set to 128. The maximum prompt and response lengths are both 4,096 tokens. We use the AdamW optimizer with a learning rate of 5×10$^{-6}$ and a weight decay of 0.01, enabling KL regularization with a coefficient of 0.01 and online filtering. Stage 2 and Stage 3 are trained for 90 and 30 steps, respectively. vLLM is used for inference and both the generation temperature and top-p set to 0.9.

For the reward mechanism and thinking strategy, the visual and textual components of the route reward are assigned weights of 30\% and 70\%, respectively, while the corresponding mixed reward weights are 10\% and 20\%. The no-vision penalty coefficient is set to 0.50, the simple-sample penalty coefficient to 0.70, the routing coefficient to 0.10, and the budget coefficient to 1.00. In the difficulty-aware fast–slow thinking mechanism, the thresholds for easy and hard samples are set to 0.80 and 0.50, respectively. The minimum thinking length is 50, and the pass threshold is set to 0.99.
\paragraph{Evaluation Setup.}
We conduct all evaluations using the VLMEvalKit framework~\citep{duan2024vlmevalkit} and locally deploy Qwen3-30B-A3B~\citep{yang2025qwen3technicalreport} as the judge model.
Our evaluation covers three groups of benchmarks:
Visual-Centric Perception, including V$^\ast$~\citep{Wu_2024_CVPR}, HRBench-4K~\citep{wang2025divide}, HRBench-8K~\citep{wang2025divide}, and MME-RealWorld-Lite~\citep{ICLR2025_df29d63a}; Reasoning-Centric Understanding, including MathVista$_{\mathrm{mini}}$~\citep{ICLR2024_663bce02}, MMMU-Pro-10-option~\citep{yue-etal-2025-mmmu}, MMMU-Pro-V~\citep{yue-etal-2025-mmmu}, and MMVet~\citep{yu2024mmvet}; and Physics-Reasoning benchmark (Out-of-Distribution), including SeePhys~\citep{NEURIPS2025_86a47273} and SeePhysPro~\citep{xiang2026seephysprodiagnosingmodality}.
\subsubsection{Datasets}
We construct two datasets for different training stages: one for SFT and the other for RL.
\paragraph{SFT dataset}

\begin{figure*}[htpb!] % 如果是双栏跨栏
    \centering
    % 左侧图 1
    \begin{subfigure}[b]{0.49\textwidth}
        \centering
        % 【关键修改】：按绝对高度缩放，而不是宽度
        \includegraphics[width=1\textwidth]{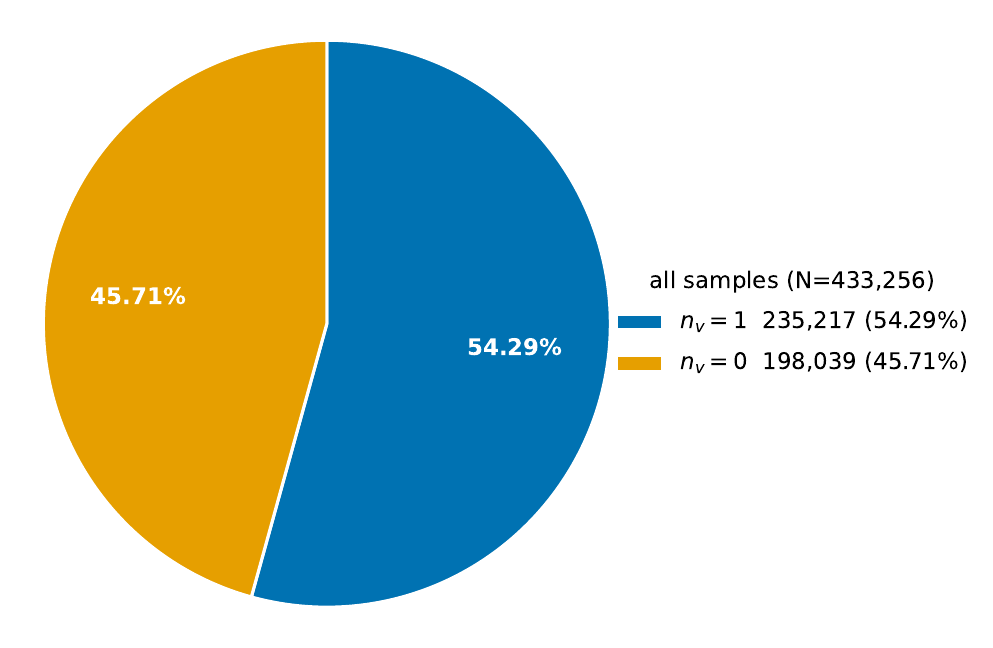} 
        \caption{Pie chart of the SFT dataset.}
        \label{sftpie}
    \end{subfigure}
    \hfill
    % 右侧图 2
    \begin{subfigure}[b]{0.49\textwidth}
        \centering
        % 【关键修改】：保持和左边完全一样的 height 数值
        \includegraphics[width=1\textwidth]{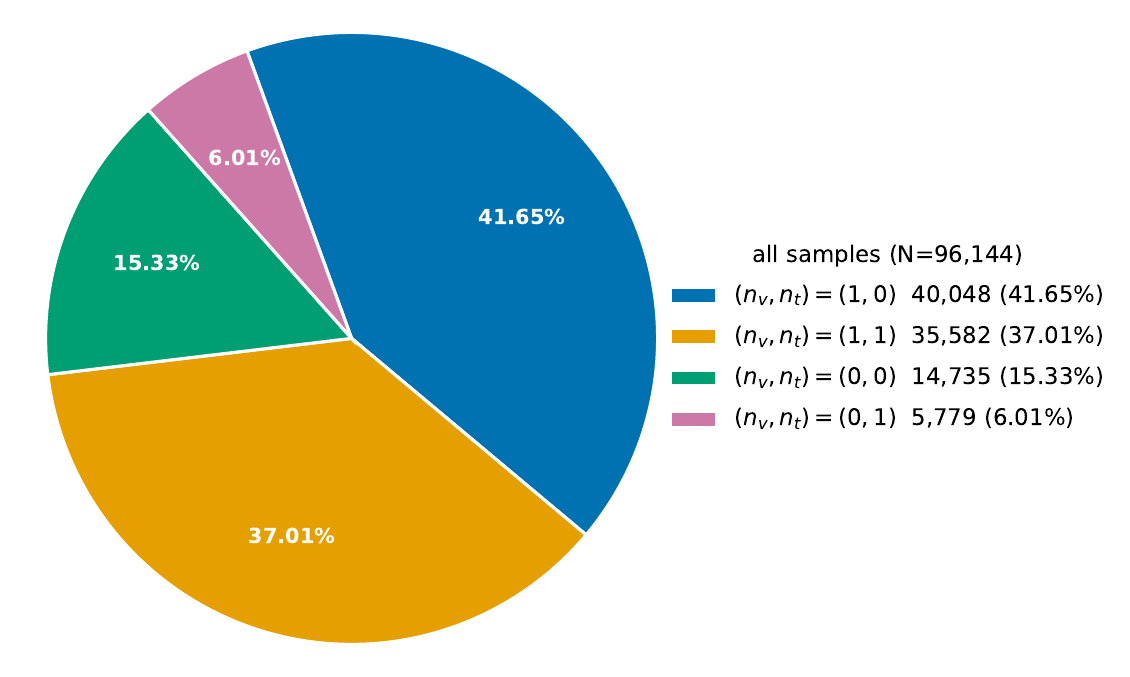} 
        \caption{Pie chart of the RL dataset.}
        \label{rlpie}
    \end{subfigure}
    
    \caption{Pie chart of the datasets.}
    \label{pie}
\end{figure*}

The SFT dataset is built from Thyme-SFT~\citep{zhang2025thyme} ('2round', 'wo thinking thyme single round'), Zebra-CoT~\citep{li2025zebracotdatasetinterleavedvision} (from Monet-SFT~\citep{wang2025monetreasoninglatentvisual}), and Mulberry~\citep{yao2026mulberry} (from Vision-R1 SFT~\citep{huang2026visionr}). We convert the intermediate visual trajectories in Thyme-RL and Zebra-CoT into discrete visual tokens using VQ-VT, while Mulberry is kept unchanged since it does not contain tool-produced visual states. As illustrated in Figure~\ref{sftpie}, we maintain an approximately balanced distribution, with around half of the training samples containing visual tokens. This design encourages the model to learn when to generate visual tokens without biasing it toward producing them for every sample.
\paragraph{RL dataset}

The RL dataset consists of Thyme-RL~\citep{zhang2025thyme}, ViRL39K~\citep{wang2026vlrethinker}, and Geo3K~\citep{lu-etal-2021-inter}. We first assign coarse routing labels according to the dataset source and the task categories provided in ViRL39K, and then further refine these labels using rule-based keyword matching. This lightweight annotation strategy provides a practical initialization for modality-aware routing without requiring expensive manual labeling.

\begin{figure*}[htbp!] % 如果是双栏跨栏
    \centering
    % 左侧图 1
    \begin{subfigure}[b]{0.49\textwidth}
        \centering
        % 【关键修改】：按绝对高度缩放，而不是宽度
        \includegraphics[width=1\textwidth]{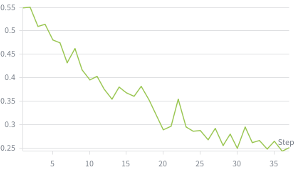} 
        \caption{Remaining ratio of informative rollout groups after online filtering during RL training on Thyme-RL.}
        \label{1}
    \end{subfigure}
    \hfill
    % 右侧图 2
    \begin{subfigure}[b]{0.49\textwidth}
        \centering
        % 【关键修改】：保持和左边完全一样的 height 数值
        \includegraphics[width=1\textwidth]{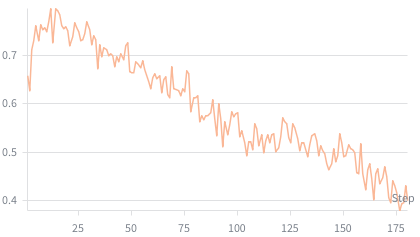} 
        \caption{Remaining ratio of informative rollout groups after online filtering during RL training on ViRL39K+Geo3k.}
        \label{f2}
    \end{subfigure}
    
    \caption{Remaining ratio of informative rollout groups after online filtering during RL training.}
    \label{fig:ratio}
\end{figure*}

As shown in Figure~\ref{rlpie}, since Thyme-RL contains substantially more samples than the other datasets, the resulting routing labels naturally exhibit a higher proportion of visual-token supervision. In addition, we observe that Thyme-RL includes many relatively simple perception tasks, which rapidly converge during RL training, we shown the result in Figure~\ref{fig:ratio}. For example, after only 30 training steps, approximately 75\% of the sampled groups are either entirely correct or entirely incorrect, leaving only 25\% that contribute effective optimization signals. In contrast, the more reasoning-oriented ViRL39K and Geo3K datasets still retain around 40\% informative groups even after 175 training steps.

To alleviate this imbalance, we employ an online filtering strategy that automatically removes fully solved perception samples from optimization. Consequently, RL updates are no longer dominated by easy perception tasks, allowing the model to focus on more informative training signals. More importantly, this prevents the policy from overemphasizing visual-token generation simply due to dataset composition, encouraging a more balanced allocation between visual and textual reasoning according to task requirements rather than sample frequency.
\subsection{limitations}
Our method still has several limitations. The visual codebook compresses intermediate visual states into discrete visual tokens, inevitably sacrificing part of the fine-grained visual information. More importantly, although visual tokens improve reasoning performance by encouraging the model to re-examine visual evidence, they cannot fundamentally correct perception errors caused by the limited representation capability of the underlying vision encoder. As demonstrated in our case studies and error analysis, once incorrect visual evidence is encoded, subsequent reasoning—whether visual or textual—has limited ability to recover from these early perception failures. This suggests that improving reasoning alone cannot fully compensate for weak visual perception, and future work should jointly enhance perceptual representation learning and reasoning.
\end{document}

%% file: math_commands.tex
\usepackage{amsmath,amsfonts,bm}

\def\eqref#1{equation~\ref{#1}}
\def\1{\bm{1}}

\DeclareMathAlphabet{\mathsfit}{\encodingdefault}{\sfdefault}{m}{sl}
\SetMathAlphabet{\mathsfit}{bold}{\encodingdefault}{\sfdefault}{bx}{n}